# Classifier Calibration: with application to threat scores in cybersecurity

Waleed A. Yousef, *Senior Member, IEEE*, Issa Traoré, *Senior Member, IEEE*, and William Briguglio, *Member, IEEE*

*Abstract*—This paper explores the calibration of a classifier output score in binary classification problems. A calibrator is a function that maps the arbitrary classifier score, of a testing observation, onto $[0, 1]$ to provide an estimate for the posterior probability of belonging to one of the two classes. Calibration is important for two reasons; first, it provides a meaningful score, that is the posterior probability; second, it puts the scores of different classifiers on the same scale for comparable interpretation. The paper presents three main contributions: (1) Introducing multi-score calibration, when more than one classifier provides a score for a single observation. (2) Introducing the exact analogy between two scenarios: (a) designing a classifier from a set of features, and (b) designing a calibrator, to generate a single calibrated score, from a set of scores of different classifiers. Hence, we propose expanding these classifiers' scores to higher dimensions to boost the calibrator's performance. (3) Conducting a massive simulation study, in the order of 24,000 experiments, that incorporates different configurations, in addition to experimenting on two real datasets from the cybersecurity domain. The results show that there is no overall winner among the different calibrators and different configurations. However, general advices for practitioners include the following: the Platt's calibrator [1], a version of the logistic regression that decreases bias for a small sample size, has a very stable and acceptable performance among all experiments; our suggested multi-score calibration provides better performance than single score calibration in the majority of experiments, including the two real datasets. In addition, expanding the scores can help in some experiments.

## I. Introduction

There are several cybersecurity models or systems (e.g. IDS, antivirus systems, vulnerability management and patching systems, and biometric authentication systems) whose output is a numerical score, which can be used for threat assessment and decision making. For instance, the common vulnerability scoring system (CVSS) is an industry standard for assessing the severity of software vulnerabilities, which produces a score ranging from 0 to 10. Another example is the OWASP risk rating scheme, which uses a score ranging from 0 to 9 to calculate an overall severity for the security risks identified in system architecture design and threat modeling. Other prominent examples are biometric technologies, which use a biometric matching score that is compared against a threshold to make crucial authentication or identification decisions. While these scoring systems can help in decision making, the underlying methodologies can be prone to subjectivity, which could diminish to some extent the reliability of the produced scores. A stronger alternative to ad-hoc scores is using probability scores for decision-making. Probability models represent a more rigorous and universally accepted measurement system. The initial motivation behind the present article was to find a systematic approach to convert the confidence scores into probability values that are ready to be used in cybersecurity models. In practice, each decision model is a classifier in its own right, and converting classifier score to a probability measure is called classifier calibration or single-score calibration. An interesting mathematical property of single-score calibration is performance invariance, that is both the true and estimated performances of a classifier are preserved after score calibration using a monotonic calibration-function, as will be detailed in Sec. III.

In addition to single-score calibration, our proposed scheme can be applied, simultaneously, to a variety of cybersecurity use cases, such as vulnerability scoring, risk rating, alert prioritization, intrusion decision making, and biometric authentication. As an example, consider a threat detection system which aggregates outputs from three different systems to determine the overall threat faced by a given host: a vulnerability monitoring system, a host-based IDS (HIDS), and an antivirus system deployed on the host. Assume that each of these systems operates independently and generates a score reflecting the severity of its respective monitoring outcomes or events (i.e., discovery of a vulnerability, intrusion alert, or virus detected, respectively). Combining the aforementioned scores, naively by normalization after addition or multiplication, can be very unwieldy and imprecise. However, through our proposed multiple-score calibration approach, the scores of the three systems can be combined and converted into a single probability measure, in a more rigorous way, producing a more reliable basis for decision-making. Although the motivation of the present article, so far, stems from the cybersecurity domain, its applicability extends to the whole machine learning (ML) community, where (multi)classifiers are needed to be calibrated.

In a more formal language, as will be detailed in Sec. II, when a black-box trained classifier tests on an observation, with a feature vector $X$ and a posterior probability $p$ (of belonging to one of the classes), it assigns a score $h(X)$ to this observation, which does not necessarily equal to $p$. When a labeled dataset of size $n$ is available for testing, the classifier can always create the set $\mathcal{H} = \{(h_i, y_i) | i = 1, \cdots, n\}$, a set of pairs of scores and true labels. Then, in single-score calibration problems, it is required to use $\mathcal{H}$ to design the

Waleed A. Yousef, Issa Traoré, and William Briguglio are with Department of ECE, University of Victoria, Victoria, BC, Canada. (email: wyousef@UVIC.ca; itraore@ece.uvic.ca; wbriguglio@uvic.ca).

Waleed A. Yousef is with Department of CS, Human Computer Interaction Laboratory (HCILab), Helwan University, Cairo, Egypt.

calibration function $\hat{p}(h)$ that maps $h$ to $p$. We observed that there is an analogy between designing a single-score calibrator and designing a classifier, because, abstractly speaking, both use a labeled set with an input (predictor) to produce an output (response). There is a further interesting analogy between a multi-score calibrator that combines different scores to produce a final calibrated score, on the one hand, and a multiple classifier system (MCS) that combines several classifiers to produce a more accurate classifier, on the other hand. This is our point of departure to provide the following contributions in the present article:

1) We introduce the multi-score calibration, to show how to provide a single calibrated probability score from multiple scores, which outperformed the single-score calibration, in terms of Brier index [2], in the vast majority of our conducted experiments.

2) We propose expanding the score(s) of the single (or multi) classifier(s) to higher dimensions, which means having several predictors in addition to the original scores, to boost the calibration accuracy of the logistic regression. This expansion can be achieved by any set of basis functions, such as polynomials, as always done for classical classification and regression problems.

3) Although a few studies (e.g., [3–5], as will be detailed in Sec. II-B) provided a comparative study for the existing single-score calibration methods, we established a benchmark for both the single- and multi-score calibration that comprises a massive number of experiments using 24,000 different configurations of simulated datasets to account for different probability distributions, sample sizes, and so forth, to be able to reach a more substantiated conclusion. All experiments were carried out on the Compute Canada Cedar Cluster[1] (CCCC), which has immense computational power that allowed us to run, on average, 500 jobs in parallel and reduced the total computational time significantly. To enable reproducing our experiments and results, and for sharing with several communities, all the code is open-sourced and made publicly available[2].

The rest of the article is organized as follows: In Sec. II we provide both a brief mathematical introduction and a literature review of the main methods of calibration. In Sec. III, we provide more mathematical formalization of the single and multi-score calibration problem, and we detail the contributions in which we extend the literature. Sec. IV is a comprehensive set of experiments on simulated datasets and on two real datasets from the cybersecurity domain. The experiments establish a systematic comparison among all known calibrators under different configurations while reproducing previous results published in [4, 5] for the sake of comparison. In Sec. V, we discuss the results more, contemplate on the discrepancy between the results of simulated data and real data, and provide a final advice for the practitioners of ML in general and network security domain in particular. Finally, Sec. VI concludes the article.

[1] https://www.computecanada.ca/
[2] https://github.com/isotlaboratory/ClassifierCalibration-Code

## II. Background and Literature Review

### A. Background

If a random vector $X$ and a random variable (r.v.) $Y$ have a joint probability density $f_{X,Y}(x,y)$, and $Y$ is qualitative (or categorical) with only two possible values (class $\omega_0$ or class $\omega_1$), the binary classification problem is defined as follows: how to find the classification rule $\hat{Y} = \eta(X)$ that can predict the class $Y$ (the response) from an observed value of the variable $X$ (the predictor). It is known that the best prediction function (classifier) that minimizes the risk is given by

$$\eta(x): \quad h(x) \underset{\omega_0}{\overset{\omega_1}{\gtrless}} th, \tag{1a}$$

$$h(x) = \log \frac{f_X(X = x|\omega_1)}{f_X(X = x|\omega_0)}. \tag{1b}$$

The log-likelihood ratio (LHR) $h(x)$ acts as the decision value, discriminant value, or the score. Now, if the joint distribution is unknown, a classification function is modeled parametrically or nonparametrically, and a training sample is used to build that model. The modeled decision function (or scoring function) $h_{\mathbf{tr}}(X)$ is built using the training dataset $\mathbf{tr}$, and the final classification rule takes the form (1a). However, of course, it is no longer the LHR nor the optimal classification rule that minimizes the risk.

The scoring function $h_{\mathbf{tr}}(X)$ is usually compared with a chosen threshold value $th$ to give the final binary decision. One common measure of assessing the decision (scoring) function $h_{\mathbf{tr}}(X)$ is the two types of error: $e_1 = \int_{-\infty}^{th} f_h(h(x)|\omega_1) \, dh(x)$ is the probability of classifying a case as belonging to class 0 when it belongs to class 1, which is called false negative fraction (FNF); $e_0 = \int_{th}^{\infty} f_h(h(x)|\omega_0) \, dh(x)$ is the opposite, called the false positive fraction (FPF). The error is not a sufficient measure, because it is a function of a single fixed threshold. A more general way to assess a classifier is provided by the receiver operating characteristic (ROC) curve. This is a plot for a true positive fraction (TPF = 1 - FNF) versus FPF under different threshold values. The area under the ROC curve (AUC) can be thought of as one summary measure for the ROC curve. Formally, $AUC = \int_0^1 TPF \, d(FPF)$, which is the same as $\Pr(h(x|\omega_0) < h(x|\omega_1))$, which expresses the separation between the two sets of decision scores $h(X|\omega_1)$ and $h(X|\omega_0)$. In practical setup, only a finite testing dataset $\mathbf{ts} = \{x_i, y_j \mid i = 1, \cdots, n_0, \ j = 1, \cdots, n_1\}$ is available to estimate the performance measures. The uniform minimum variance unbiased estimators (UMVUE) for the AUC, under the nonparametric assumptions, is given by the Mann-Whitney statistic [6–9]

$$\widehat{AUC} = \frac{1}{n_1 n_0} \sum_{i=1}^{n_0} \sum_{j=1}^{n_1} \psi\big(h(x_i|\omega_1), h(x_j|\omega_0)\big), \tag{2a}$$

$$\psi(a,b) = \begin{cases} 1 & a > b \\ 1/2 & a = b \\ 0 & a < b \end{cases} \tag{2b}$$

In many applications, it is always desirable that the classifier output score corresponds to the posterior probability $\Pr(\omega_i|X = x)$, $i = 0, 1$. Suppose that we have two classifiers giving scores for an observation of 3.5 and 17.4, respectively.

It is not only true that each score is non-informative but also both scores cannot be combined in a direct way because they are on two different scales akin to the particular design process of each classifier. To wit, for example, in cybersecurity, the score of a classifier—either an intrusion detection system (IDS) or anti virus (AV)—should correspond directly to the probability of, say, being malicious ($\omega_1$) given the feature vector $X = x$. This will allow for the interpretation and comparison of different scores. The connection between the posterior probability and score likelihood is straightforward through Bayes' theorem, as follows:

$$\frac{\Pr(\omega_1|h)}{\Pr(\omega_0|h)} = \frac{f_h(h|\omega_1)\Pr(\omega_1)/\Pr(h)}{f_h(h|\omega_0)\Pr(\omega_0)/\Pr(h)}, \tag{3}$$

After solving for $\Pr(\omega_1|h)$, giving it a shorthand notation of $p$, we get:

$$p(h) = \frac{1}{1 + L^{-1}(1-\pi)/\pi}, \tag{4a}$$

$$L = f_h(h|\omega_1)/f_h(h|\omega_0), \tag{4b}$$

$$\pi = \Pr(\omega_1). \tag{4c}$$

An aside technical point to mention is that the probability $p(h)$ in (4a) is the posterior probability $\Pr(\omega = \omega_1|h)$, given a score value $h$, not the posterior probability $\Pr(\omega = \omega_1|X = x)$ given a feature vector $X = x$. This may only be important for theoretical treatment in the cases that the score function $h(X)$ assigns the same value to two or more vectors $X$.

The problem of classifier calibration (surveyed in Sec. II-B) is designing the estimator $\widehat{p}(h)$ as "close" as possible to the true value $p(h)$ in Eq. (4). Because $p \in [0,1]$ is no longer an indicator function, as in the original classification problem, the two common accuracy measures used in the literature are the mean square error (MSE) and the Brier score, which are defined as:

$$\text{MSE}(\widehat{p}, p) = \mathop{\mathrm{E}}_{h}\big[\widehat{p}(h) - p(h)\big]^2, \tag{5a}$$

$$\text{Brier}(\widehat{p}, p) = \mathop{\mathrm{E}}_{h}\big[\widehat{p}(h) - y\big]^2, \tag{5b}$$

where $y(=0,1)$ is the true class label of the observation with score $h$. However, we always prefer the root version of each measure, $\text{RMSE}(\widehat{p}, p) = \sqrt{\text{MSE}(\widehat{p}, p)}$, $\text{RB}(\widehat{p}, p) = \sqrt{\text{Brier}(\widehat{p}, p)}$ because it has the same units, not a square, of the estimand. For a black-box trained classifier, and with no knowledge of the probability distributions of scores, at most we can have an access to a labeled dataset, from which we can construct the set

$$\mathcal{H} = \big\{(h_i, y_i) \mid i = 1, \cdots, n\big\}, \tag{6}$$

where $h_i$ is the score of the $i^{\text{th}}$ observation and $y_i$ is its label. Then, calibration performance measures, can be estimated by:

$$\widehat{\text{RMSE}}(\widehat{p}, p) = \sqrt{\frac{1}{n}\sum_i (\widehat{p}(h_i) - p(h_i))^2}, \tag{7a}$$

$$\widehat{\text{RB}}(\widehat{p}, p) = \sqrt{\frac{1}{n}\sum_i (\widehat{p}(h_i) - y_i)^2}. \tag{7b}$$

However, because $p(h_i)$ requires knowledge of the parametric distributions (per Eq. (4)), the estimate (7a) is of practical use only for simulated datasets, and the estimate (7b) is suitable for both simulated and real datasets. However, this will be a trade-off between what we need, i.e. the error $\widehat{p}(h_i) - p(h_i)$ and what we have, that is the error $\widehat{p}(h_i) - y_i$.

As was introduced above, the following observation is quite interesting and is the basis for Sec. III, which is an extension of the literature on calibration. From a pure mathematical point of view, the function $\widehat{p}(h)$ that calibrates the classifier output $h(X)$, that is obtained from the feature vector $X$, is itself a classifier that trains on the dataset (6) with a single feature $h(X)$ and output score function $\widehat{p}(h)$. In the sequel and to avoid any confusion of mixing up terminology, we will exclusively reserve the terms classifier or score/scoring function to mean $h(X)$ and will be using them exchangeably as needed by context. On the other hand, we will use the term calibrator to mean $\widehat{p}$, which converts the initial score(s) to a calibrated probability score.

### B. Literature Review

Before reviewing the literature of the single-score calibration, we first ponder equation (4). Modeling the calibration function $p(h) : \mathbf{R} \mapsto [0,1]$ can be accomplished via several approaches. We provide our view of these general approaches and survey the literature belonging to each of them. We emphasize that our calibration problem deals with a black-boxed classifier with no access to their internal structure and only the classifier score along with true labels, that is the dataset (6), are available. This should be distinguished from other approaches, e.g., [10], where they assume some knowledge of the internal basis kernel of the classifier and use the labeled dataset to estimate some of their parameters.

*1) Modeling the Likelihood (LH):* In this approach, the two LH functions $f_i(h|\omega_i)$, $i = 0, 1$ are modeled as a first step. This is nothing but an estimation of a univariate PDF function, which belongs to rich statistical literature that includes both parametric and nonparametric methods. After this first step, the posterior probability (4a) is modeled by plugging in the estimated LHs $\widehat{f}_i(h|\omega_i)$ along with $\widehat{\pi} = n_1/n$. An example of the literature leveraging this approach is [11], where the estimation of the LH is done by the basic binning method (the same method used for drawing histograms) by treating the bin width as a tuning parameter, which can be obtained via cross validation (CV). Therefore, the LH estimation is given by the relative number of observations belonging to a bin interval $B$. Formally, for $x$ belonging to a bin interval $B$, then

$$\widehat{f}_i(x|\omega_i) = \frac{1}{n_i}\sum_{j=1}^{n_i} I_{x_j \in B}, \ i = 0, 1. \tag{8}$$

It is clear that the resulting calibration function will be piece-wise constant (within the interval of each bin), with discontinuities (jumps) at the boundaries of each bin. Therefore, histogram binning calibration produces an unsmooth function, because it is not differentiable at these points.

*2) Modeling the Likelihood Ratio (LHR):* In this approach, the LHR (4b) (not the individual LH functions) is modeled;

| Method | Approach | Availability | Smoothness | Multi-score | Performance |
|--------|----------|--------------|------------|-------------|-------------|
| logistic regression [1] | PP | 🟢 | 🟢 | 🟢 | 🟢 |
| histogram binning [11] | LH | 🟢 | 🔴 | 🔴 | 🟡 |
| isotonic regression [20] | PP | 🟢 | 🟡 | 🔴 | 🟡 |
| PROPROC [4, 5] | LHR | 🔴 | 🟢 | 🔴 | 🟢 |

Table I: Summary of score calibration methods (with rows ordered chronologically by the date of the cited references). For each method (row), we list its modeling approach (likelihood (LH), likelihood ratio (LHR), or posterior probability (PP)) and whether it is good (🟢), fair (🟡), or poor (🔴) for each of the four aspects: availability in scientific software, smoothness of the calibrated score curve, support of multi-score calibration, and performance in comparative studies. Performance is measured in terms of RMSE and RB (7) in our present study and other literature studies.

then, similar to above, this estimate, along with the estimate of the prior probability, are plugged in (4a). There is, as well, a wide range of statistical literature for modeling the LHR of a two sample data. However, the literature is not leveraged in the score calibration problem except by [4, 5], where the authors used PROPROC software for fitting the ROC of a two-sample data to estimate the LHR (see [12]). They call this method the semi-parametric approach because the parametric model used involves some internal latent variables. The details are out of the scope of the present article and can be found in the literature on ROC modeling of two-sample data that were well developed and established by Metz *et al.* during the previous two decades, for example in [12–19]. By construction, the parametric model of the PROPROC is differentiable everywhere, and hence the calibration function is smooth.

*3) Modeling the Posterior Probability (PP):* In this approach, the posterior probability (4a) is directly modeled. This is a typical regression problem that can be done parametrically or nonparametrically. [20] seems to be the first to propose the nonparametric regression for modeling the posterior probability directly from score values. The authors proposed using the isotonic regression, an optimization method that minimizes the square loss under the constraint that the resulting regression function is monotonic (not strictly monotonic, as will be distinguished in definition (1)). Therefore, the method provides an unsmooth regression curve with piecewise constant segments, exactly as the histogram binning does. [1] suggested using logistic regression to model the posterior probability from scores. Their initial motivation was to model the score of the SVM in particular. The modeled regression function is the typical smooth sigmoid function, which is differentiable everywhere. However, they used a slightly modified version of the known logistic regression to decrease the bias for a small sample size. It is worth mentioning that [10] follows [1] but with an additional ingredient of estimating some parameters related to the basis kernel of the classifier. As was introduced above, this is a bit different from our assumption of knowing nothing about the internal structure of the classifier.

*4) Comparison:*

It is pedagogical to put the single-score calibration methods, reviewed above, side by side in a comparison and comment on each, before introducing both the multi-score and score-expansion calibrations. We order these methods chronologically in Table I, which indicates the modeling approach (LH, LHR, or PP) of each. All methods, except PROPROC, are available in almost any scientific computing environment (SCE). Although the code of the PROPROC software is publicly available and can be integrated to SCEs, the version that provides an estimate of the LHR, which is necessary for calculating the calibrated scores, is not publicly available. Only logistic regression and PROPROC introduce smooth calibrated curves, as opposed to histogram binning. The isotonic regression produces a step-wise calibrated curve, which is not smooth. However, because the intervals of these steps are usually, by construction, shorter than the histogram bins, isotonic regression is less rough (less unsmooth) than the histogram binning. Only the logistic regression is ready for multi-score calibration, because it can accept a feature vector of any dimension. The histogram binning is ready for binary-score calibration, because 2D histogram binning exists in all SCEs. Although higher-dimension histogram binning can be trivially programmed by creating a $d$-dimensional grid, many bins will be empty, casting the method with no practical utility. Both isotonic regression and PROPROC require theoretical development to apply multi-score calibration. There are a few studies, for example [3–5], that compare the performance of these single-score calibration methods. These studies assumed different score distributions either by explicitly specifying the distribution and playing with its parameters or producing the score from trained classifiers on different datasets. Their conclusion is that there is no overall winner over all the score distributions; however, the logistic fit [1] and the semiparametric fit [5] approaches are very acceptable. We reached at a somewhat similar conclusion (more details are provided in Sections IV and V); however the massive number of experiments that we conducted (24,000 different configurations, including probability distributions, sample sizes, etc.) provides a solid ground and substantiation for our conclusion. Based on the four-criterion comparison above, the only candidate for both the multi-score and the score-expansion calibrations is the logistic regression, as will be detailed in the next section.

### C. Introductory Pedagogical Example on Calibration

It is more informative to illustrate the calibration process on some dataset. We use the labeled subset of one of the datasets from the 2019 IEEE BigData Cup Challenges. The dataset consists of event log data received at a security operation center (SOC) run by a company called Security On-Demand (SOD) [21]. The considered subset contains 39,427 observations. The single feature `overallseverity`, which represents the intrusion alert severity generated by the system rules, is an integer value $0, 1, \cdots, 5$, which accounts for the score of an intrusion detection system (IDS). The column `notified` contains the true label to indicate whether the observation is an intrusion or a normal behavior. The 39,427 records include 2,276 intrusions and 37,151 normal records.

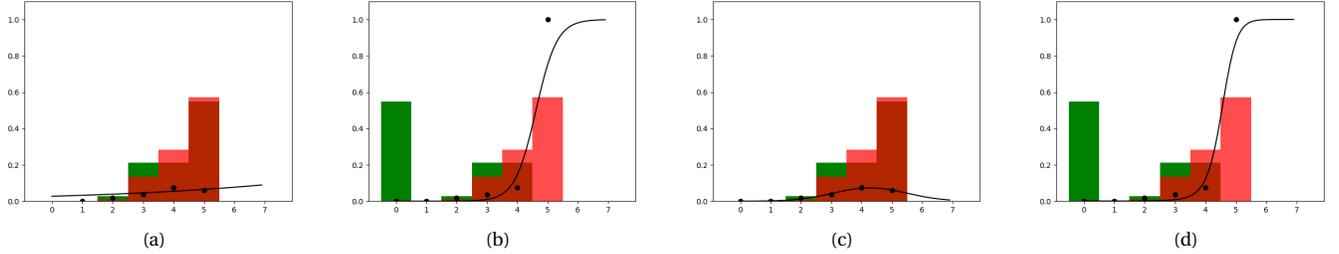

Figure 1: Histograms binning and logistic regression calibration of the scores of an IDS system. (a) without score modification, where the two sets of score are almost overlapping and non-predictive (Mann-Whitney of 0.53). The histogram fitting is the black dots (very unsmooth) the logistic fit is the sold line (very smooth). (b) after artificial score modification, where more natural separability is obvious (Mann-Whitney = 0.92). (c)–(d) the same experiment but with introducing two features, both the score and its squared values (Sec. III-B2), to the logistic regression model. The fit is obviously better.

Because the dataset size is very large, the nonparametric estimate of the LHR, and hence, the posterior probabilities, can be considered a very accurate estimate for the population values.

Before showing how to train some of the calibrators using the techniques discussed in Sec. II-B, it is important to comment on this dataset. The score `overallseverity` is not a discriminating feature. The Mann-Whitney statistic (2a) of this score is 0.53, which is almost close to a random guess. Just for the sake of clarification, modifying the value of 5 of the score of the normal records to 0 (to mimic a more realistic and practical IDS scoring) gives a Mann-Whitney statistic of 0.92. Figure 1 illustrates the histograms of the two classes for the data before, and after, modification.

We calibrate this score before and after modification using two calibration methods: the smooth logistic regression and the very unsmooth histogram binning method. Calibrating the original score gives the poor model of Figure 1a, where the binning method is plotted as individual dot markers and the logistic fit is plotted as a solid curve. (The solid curve is just for illustrating the smoothness of the method; however, because the score only takes the discrete values $0, \cdots, 5$, the curve is discrete as well). This poor fit rises from the fact that the two classes almost have the same likelihood functions, which is obvious from both the visual histogram and from the Mann-Whitney value of 0.53. For example, at a score value of 5, the two likelihoods are almost the same, so the posterior probability (4a) will be almost the prior probability $\pi$ estimated by $n_1/n = 2276/39427 = 0.058$, which is almost the same value provided by the logistic fit.

Calibrating the modified score produces the reasonable calibration function of Figure 1b. However, the logistic fit at the score value of 5 is not very accurate: $\widehat{f}_1^{logit}(h = 5|\omega = \omega_1) = 0.75$; whereas, the binning method gives an estimate of exactly 1 because $f_h(h = 5|\omega_0) = 0$, which gives an estimate of the LHR of $\infty$. When we provide the logistic regression with each value of the score $h$ and its square $h^2$ as an expanded feature, (this approach is detailed in Sec. III-B), the fit to the posterior probabilities is better, as in Figures 1c–1d.

## III. Score Calibration: formalization

Given a classifier that provides a scoring function $h(X)$, the calibration function is defined as a function that maps $h(X = x)$ to $\Pr(\omega_1|h)$. This mapping is called the calibration function. In this section, we first formalize the calibration problem in more mathematical rigor (Sec. III-A); then, we propose how to calibrate multiple scores, and how to expand the available score to higher dimensions to boost the calibration accuracy of both single and multiple scores (Sec. III-B). We also discuss the attempts of calibrating multiple scores without having access to labeled datasets (Sec. III-C).

### A. Single Score Calibration

**Definition 1.** *[Calibration Function] is a monotonically increasing mapping from* $\mathbf{R}$ *to* $[0, 1]$ *to map a classifier output score to a posterior probability.*

This formal definition verbally means the following: Consider the two observations $x_1$ and $x_2$, belonging to any set (not necessarily $\mathbf{R}$) along with their classifier output scores $h(x_1) < h(x_2)$. The calibration function (call it $p(h)$) that estimates the posterior probabilities should satisfy $p(h_1) < p(h_2)$ for a strictly monotonic assumption and $p(x_1) \leq p(x_2)$ for a monotonic assumption. This distinction is being formally made here to conform with some methods in the literature and is discussed in Sec. II-B. The ROC and AUC are invariant under the strictly monotonic calibration (Corollary 3). This means that both the original classifier score or its calibrated score have the same ROC and classification accuracy at two equivalent threshold values as stated by the following two lemmas.

**Lemma 2.** *Consider any two r.v.* $X$ *and* $Y$ *belonging to* $\mathbf{R}$ *with distribution functions* $F_X$ *and* $F_Y$, *respectively, along with a strictly monotonically increasing function* $p$, *and* $p(X) = X' \sim F_{X'}$ *and* $p(Y) = Y' \sim F_{Y'}$. *Then:*

1) *For every threshold value* $\theta \in \mathbf{R}$, *with the corresponding probabilities* $F_X(\theta)$ *and* $F_Y(\theta)$, *there exists a unique corresponding threshold value* $\theta' = p(\theta)$, *with the corresponding probabilities* $F_{X'}(\theta')$ *and* $F_{Y'}(\theta')$ *such that* $F_{X'}(\theta') = F_X(\theta)$ *and* $F_{Y'}(\theta') = F_Y(\theta)$.

2) *For every threshold value $\theta' \in \mathbf{R}$, the converse argument applies.*

3) *These two properties are not satisfied with relaxing the strictly monotonic assumption and considering $p$ a monotonic increasing function.*

**Proof.** Each case is proven separately:

1) Because the mapping $p$ is strictly monotonically increasing, it is a bijection. Therefore, $\forall x \in [-\infty, \theta]$ there exists a unique $x' = p(x) \in [-\infty, \theta]$, where $\theta' = p(\theta)$; and therefore, the two sets are equivalent. Hence, $F_X(\theta) = \Pr[X \le \theta] = \Pr[X' \le \theta'] = F_{X'}(\theta')$. The same argument is immediate for $Y$.

2) Because the function $p$ is a bijection, its inverse $p^{-1}$ is as well; then, the argument is identical to the one above.

3) Consider a non-strictly monotonically increasing function $p$ such that $p(x_1) \le p(x_2)$ $\forall x_1 < x_2$. Take any $\theta_1 < x_2$, where $p(\theta_1) = p(\theta_2) = \theta'$. Then, for any $\theta_1 < \theta < \theta_2$, we have $p(\theta) = p(\theta')$. Suppose that for some $\theta \in ]\theta_1, \theta_2[$, $\Pr[X \le \theta] = \Pr[X' \le \theta']$. However, of course, $\Pr[X \le \theta] \ne \Pr[X' \le \theta']$ $\forall \theta \in ]\theta_1, \theta_2[$ unless $\Pr[\theta < X \le \theta_2] = 0$, which is not necessary for any distribution $F_X$. $\blacksquare$

**Lemma 3.** *Consider applying a strictly monotonic calibration function to the output score of a classifier, and then treating its output as the new final score. Then, all the performance measures defined in Sec. I, that is errors, accuracy, risk, ROC, and AUC, along with their estimators, are invariant under this transformation.*

**Proof.** is immediate by direct application of Lemma 2. $\blacksquare$

### B. Multi-Score Calibration

Suppose we have access to more than one classifier instead of just a single classifier. Then, much like being able to generate the dataset (6), we can generate the following dataset:

$$\mathcal{H}_{1,2} = \Big\{ \big(h_1(x_i), h_2(x_i), y_i\big) \mid y_i = \omega_1, \omega_2, \ i = 1, \cdots, n \Big\}. \quad (9)$$

This dataset can be used to produce a calibrator in either a direct approach or in a training and testing approach. This should boost both the classification and calibration accuracies of the resulting calibrator. Those two approaches are elaborated on in the following two subsections. Although the material presented here treats combining only two classification scores, it is immediately extensible to $K \ge 2$ classifiers by providing the general set $\mathcal{H}_{1, \cdots, K}$, and extension to (9).

*1) Direct approach:* The scores $h_1(x), h_2(x)$ for the dataset $\mathcal{H}_{1,2}$ have a 2D PDF. Formally, and similar to (4), we can relate the posterior probability to this joint PDF by:

$$\Pr(\omega_1 \mid h_1, h_2) = \frac{1}{1 + L^{-1}(1 - \pi)/\pi}, \quad (10)$$

$$L = \frac{f_{h_1, h_2}(h_1, h_2 \mid \omega_1)}{f_{h_1, h_2}(h_1, h_2 \mid \omega_0)},$$

$$\pi = \Pr(\omega_1).$$

Therefore, in principle, the three approaches of calibrating a single classifier (LH, LHR, and PP in Sec. II-B) are immediate. However, implementing the PP approach via logistic regression, is favored because of all of the reasons discussed in the comparison provided in Sec. II-B4 and summarized in Table I.

| $\omega$ | $h_1$ | $h_2$ | $\wedge$ | $\vee$ |
|---|---|---|---|---|
| 0 | 0 | 0 | 0 | 0 |
| 0 | 0 | 1 | 0 | 1 |
| 0 | 1 | 0 | 0 | 1 |
| 0 | 1 | 1 | 1 | 1 |
| 1 | 0 | 0 | 0 | 0 |
| 1 | 0 | 1 | 0 | 1 |
| 1 | 1 | 0 | 0 | 1 |
| 1 | 1 | 1 | 1 | 1 |

Table II: Truth table of the eight possible combinations of the true label of an observation along with the corresponding output of two binary classifiers. The last two columns of the table are the simple AND and OR classifiers combination. None of the very aggressive AND rule or the very loose OR rule always improves the final classification decision; it depends on when the two classifiers agree/disagree, that is the joint PDF of both classifiers.

*2) Elaborate approach to Boost Performance:* We can treat the calibration problem as—and indeed it is—a learning problem where the training dataset is the labeled two-feature dataset $\mathcal{H}_{1,2}$ and the model is logistic regression. Therefore, we do not have to restrict ourselves to the two predictors $h_1$ and $h_2$; rather, we can expand them to a higher dimensional space by including, for example, $h_1^2$, $h_2^2$, $h_1 h_2$, in addition to the two original features. This feature expansion is a mathematical transformation from $\mathbf{R}^2$ (the space of the original two features) to $\mathbf{R}^5$ (the space of all five features). As well known in ML this transformation can be, in principle, to any new space $\mathbf{R}^p$, and by using any basis functions, polynomials or others. This expansion is necessary because the best classification function that expresses the relationship between the class label (the response) and the original features (the predictors) needs not to be linear. The optimal $p$ is chosen by CV, stage-wise forward regression, best subset selection, or regularization, among many others readily available techniques from the literature on ML. In this elaborate approach, not only will the output be calibrated to the appropriate posterior probability (10), but also the overall classification accuracy will be better than that achieved by each individual feature (classifier output) $h_i$.

### C. Combining Scores Without Labeled Dataset

There are many attempts to combine classifiers from an unlabeled dataset. Consider the existence of two black-box trained classifiers, along with their accuracies, without the availability of a labeled dataset. Therefore, the set $\mathcal{H}_{1,2}$ defined in (9) is not available as well. What is available is only the pairs $(h_1(x_i), h_2(x_i))$ $\forall i$, which are produced by direct testing of the classifiers on the observations $x_i$. Before investigating how the two scores can be combined, which has been done (as will be discussed in Sec. III-C2), we need to discuss some important aspects.

*1) Known But Usually Ignored Facts:*

Indeed, $h_1$ and $h_2$ are not only two dependent random variables (r.v.s), but they are also correlated because they are designed to produce high (low) scores for malicious (normal) activities. It should be clear that the higher their accuracies, the more their correlation because the majority of the time they will agree on the right decision, a consequence of high accuracies. Combining these scores in general and estimating the posterior in particular is theoretically impossible without any further assumptions about their joint PDF. This is clear from the posterior expression (10). For more intuition, let us consider the counter example depicted in Table II. The table represents the truth table of the binary class $\omega$ of an observation $x$ along with the outputs $h_1(x)$ and $h_2(x)$ of two binary classifiers. It is clear that the simple AND rule wins over the simple OR rule in two cases (the green cases), whereas the situation is reversed in the other two cases. In four cases, when the classifier outputs agree, the two rules have no effect on the final decision. However, under some assumptions for the classifiers' output and their dependence/correlation, some rules can prove superior over other rules. This is known, and we tersely summarize it Sec. III-C2.

*2) Methods Existing in Literature:* Combining the output of two classifiers, whether each output is a posterior or a mere score, and where there is no labeled dataset as $\mathcal{H}_{1,2}$ in (9), was studied more than 20 years ago. It started as a very ad-hoc field by suggesting naive combination rules, for example sum, product, and so forth. There was no mathematical justification for the underlying assumptions under which these rules should work. In addition, the probabilistic dependence between the outputs of the classifiers (explained above) was completely ignored. As Hastie et al. put it, *"there is a vast and varied literature often referred to as "combining classifiers" which abounds in ad-hoc schemes for mixing methods of different types to achieve better performance. For a principled approach, see Kittler et. al. (1998)."* [22, p. 624]. In [23], they justified the mathematical assumptions under which the simple combination rules should work. Afterwards, [24] derived the errors of these combination rules for some particular probability distributions.

Knowing only the accuracies of the classifiers and with zero knowledge and no assumptions of the joint density function of the classifier outputs, we do not see any mathematical justification for combining the output scores except from a simple Bayesian point of view. This is pursued by assigning a subjective prior for each classifier proportional to its relative accuracy. Said differently, the accuracy in this context plays the role of a subjective belief (or trust) that justifies taking its output seriously. The mathematical argument is explained in [22, Sec. 8.8] and goes as follows: suppose each classifier $\eta_k$, $k = 1, \cdots, K$ is already calibrated and produces the output $\Pr(\omega_1 | x, \eta_k)$ for an observation $x$, then

$$\Pr(\omega_1 | x) = \sum_k \Pr(\omega_1 | x, \eta_k) \Pr(\eta_k | x), \tag{11a}$$

$$\Pr(\eta_k | x) = \frac{A_k}{\sum_{k'} A_{k'}}, \tag{11b}$$

where $\Pr(\eta_k | x)$ is the subjective prior assigned to the $k^{\text{th}}$ classifier and $A_k$ is its accuracy. However, the subjective priors (11b) can be any other function that is monotonic in accuracy and sum up to 1 (e.g., [25] raise accuracies to an arbitrary power).

## IV. Experiments and Results

In this section, we conduct some experiments using simulated datasets and two popular public datasets from the cybersecurity domain.

### A. Simulated Datasets

*1) Single Score Calibration:* We leverage the generalized lambda distributions (GLD), which was introduced in [26], to generate a wide variety of score distributions. We follow [27, 28] in using the GLD but in a more systematic way. The GLD is a univariate unimodal distribution family with a set of four population parameters; we use the notation $\mathcal{L}(\lambda_1, \lambda_2, \lambda_3, \lambda_4)$ to denote such a distribution. These parameters determine the characteristics and shape of the distribution, for example location, skewness, etc. By adjusting these population parameters $\lambda_i$, $i = 1, \cdots, 4$, we can generate three basic distributions that are symmetric, skewed left, and skewed right. In addition, we add the normal distribution $\mathcal{N}(\mu, \sigma)$. For the score calibration problem, we have two distributions $F_0$ and $F_1$; each can be chosen from this list of four distributions, which results in 16 pairs. For each of these 16 pairs, without loss of generality, we fix the mean of $F_0$ to zero; we set the standard deviation to 1; and we adjust the location parameter of $F_2$ to obtain a desired separability (measured in AUC) between the two distributions; namely, we choose AUC = 0.6, 0.75, 0.9. This is done to express the different discriminating power of a particular classifier. Figure 2 lists the parameters of the basic four distributions and illustrates the 16 pairs; each pair is illustrated by three AUC values. Each of these 16 pairs is given an index of $1, \cdots, 16$, which is the same index of the $x$-axis of Figure 3a–3d, which will be explained later. This index is not only a mere ID assigned to each pair; rather, it has an important ordering significance that will be explained when explaining that figure. A minor technical issue is that, because of the mathematical formalization of the GLD, the parameter values shown in Figure 2 do not give a distribution directly with $\sigma = 1$. It is quite tedious to find the closed from of $\sigma$ and solve for $\lambda_i$. Rather, we simulate a large sample and divide by the estimated standard deviation. The smoothed histogram of this standardized version is what is plotted in the figure and is what will be used for further simulations. It is remarkable that to plot these curves with that smoothness, especially for the heavy tailed pairs, it required simulating up to a million observations.

For each of these $16 \times 3$ pairs, we studied the behavior of different calibrators under different balanced class sample sizes of $n = 10 \times 2^i$, $i = 0, \cdots, 9$. A single configuration (or experiment) is described as follows: We select a pair out of the $16 \times 3$ pairs, simulate a training dataset of $n$ scores for each class, train each calibrator on this training dataset, and **a** $\mathcal{L}(\lambda_1,\text{-}0.1125,\text{-}0.1359,\text{-}0.1359)$     **b** $\mathcal{L}(\lambda_1,0.014,0.009695,0.0285)$     **c** $\mathcal{L}(\lambda_1,0.014,0.0285,0.009695)$     **d** $\mathcal{N}(\mu,1)$

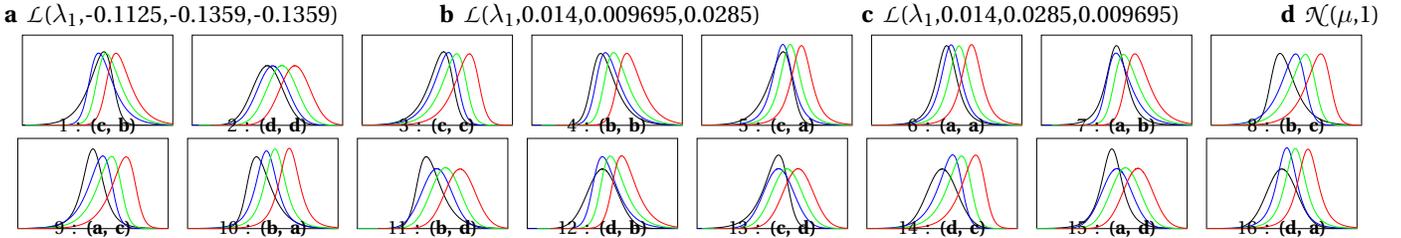

Figure 2: The parameters of the basic 4 distributions {**a**,**b**,**c**,**d**} for GLD symmetric, GLD left skewed, GLD right skewed, and Normal, respectively; and the 16 plots of their cross product used for single-score calibration. Each plot is labeled as $x:(\alpha,\beta)$, $x=1,\cdots,16$; $\alpha,\beta=$ **a,b,c,d**, were $x$ is its index used in plotting Figure 3. $\alpha$ is distribution $F_0$ with location parameter ($\lambda_1$ or $\mu$) set to zero (plotted in black). $\beta$ is distribution $F_1$ with location parameter set to three different values to give a separation with $F_0$ of AUC = 0.65, 0.7, 0.9 (plotted in blue, green, and red, respectively on each of the 16 plots). The Figure, therefore, illustrate $16 \times 3$ configurations.

finally test and assess each calibrator on: (a) the same training dataset to get both $\text{RMSE}_m^{sub}$ and $\text{RB}_m^{sub}$ (resubstitution) and (b) test them on pseudo-infinite testing set of 10,000 scores generated form the same distribution pairs to get $\text{RMSE}_m^{ind}$ and $\text{RB}_m^{ind}$ (independent). The subscript $m$ denotes the $m^{\text{th}}$ Monte-Carlo (MC) trial, which will be repeated $M$ times to obtain an average: $\text{E}_M(\cdot) = \frac{1}{M}\sum_m(\cdot)$. This is to account for the variability of the training datasets of finite size $n$.

The calibrators used in each configuration are those discussed in Sec. II-B, excluding the PROPROC for reasons discussed in Sec. IV-A2. These calibrators are as follows:

**Logistic Regression (LogReg)** with three different versions: (1) Platt's version [1], which we implement in Python by literal translation (line by line) from Matlab code written and shared to us by the authors of [4, 5]—this can then be compared to their results; (2) the `scikit` implementation [29]; and (3) the `scikit` implementation on the expanded score space as we suggest in Sec. III-B2. For short, in the sequel, we call these three versions Platt, LogReg, and LogRegExt.

**Isotonic Regression (IsoReg)** of the `scikit` package [29].

**Binning Method (BinReg),** which we implement by using five different values of bins: $10, 20, \cdots, 50$.

Therefore, we have nine different calibrators to run on each configuration. To summarize, we have a total of 480 experiments ($16 \times 3 \times 10$), and each is repeated 1000 MC trials ($M = 1000$) for assessing nine calibrators, to finally produce four error measures: RMSE and RB on both independent and resubstitution testers.

The results of these experiments are illustrated in Figure 3, which consists of five subfigures (a)–(e) (subfigures (a)–(d) are deferred to the supplementary material). Subfigures (a)–(b) illustrate the RMSE as a matrix of plots; each plot shows both $\text{RMSE}^{ind}$ and $\text{RMSE}^{sub}$ for each of the nine calibrators versus the index of the distribution pair explained in Figure 2. There are 30 plots corresponding to the $3 \times 10$ cross product of the AUC and $n$ values. Similarly, subfigures (c)–(d) (supplementary material) illustrate RB. The index of the distribution pair on the $x$-axis is not just an ID for the pair; rather, it expresses its rank for the average $\text{RMSE}^{ind}$. This average is calculated over the nine calibrators across the $3 \times 10$ configurations. Therefore, the *overall* trend of the curves is increasing with that index. This is done to visualize the effect of the distribution pair on the performance of calibrators in a more informative way. Subfigure (e) illustrates the average of each of the four metrics versus the sample size $n$. This average is taken, conditionally on each $n$ over all the $16 \times 3$ configurations for each calibrator. A common legend of subfigures (a)–(e) is displayed below subfigure (e). For the best visualization experience and figure interpretation, all plots are produced with LaTeX to control their position at the pixel level so that when pages scroll, all fixed information, for example axis, ticks, and so forth, look stationary and only components carrying information, for example curves, run as if they are animated.

Before delving into the details of this figure, it is important to study its summary from subfigure (e). It is obvious that at a smaller $n$, the two versions of logistic regression, LogReg and Platt, behaves better than all others, including the third version LogRegExt. IsoReg comes second to LogReg. Then, all calibrators improve and converge almost to the same value, as $n$ increases, with a slight superiority of the non-parametric BinReg. This comes as no surprise because it is known that the histogram method converges pointwise to the PDF of a r.v. Also, we can observe that a smaller number of bins works better at smaller sample size $n$, and vice versa, quite as expected. This summary subfigure is the final message and of practical value, to practitioners to whom the only available information in the calibration problem is the sample size $n$.

We can infer more about the calibrator's behavior by conducting a deeper investigation of 3a–3d. A pilot view of subfigures 3a–3b reveals that there is no calibrator that is a consistent winner across all configurations. In addition, the majority of calibrators, with a few exceptions, converge to the same value as $n$ increases. It is obvious that the larger the AUC, i.e., the more the classifier is capable of separating the two classes, the smaller the calibration errors. LogRegExt slightly wins across all distribution pairs only at small values of both $n$ and the AUC. It deteriorates at higher AUC values, in contrast to almost all other calibrators. This may be interpreted as follows: at a higher AUC, the two distributions are already separated, and the problem is much easier for LogReg; indeed, it is not worth the extra variance component

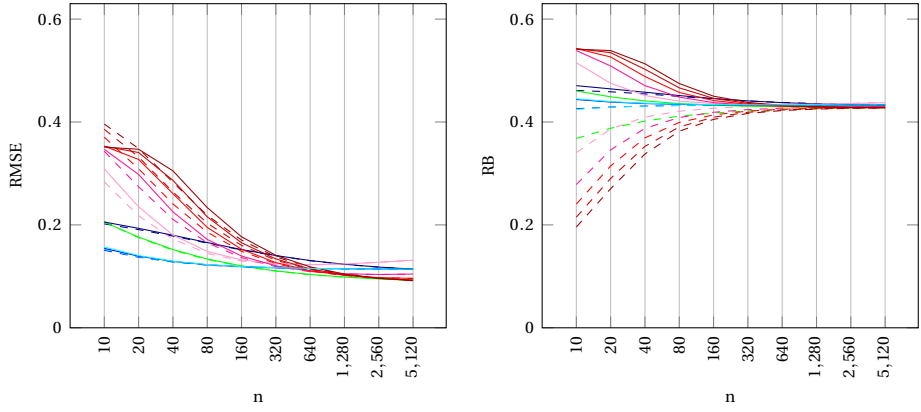

Figure 3e: single-score calibration: summary plot of (a)–(d).

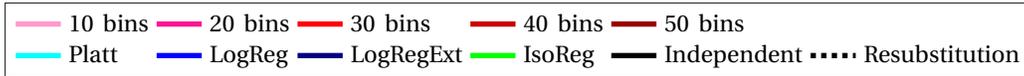

Figure 3: consists of 5 subfigures (a)–(e) with common legend for calibrators. (a)–(b) illustrate RMSE$^{ind}$ and RMSE$^{sub}$ of the 9 calibrators for the $16 \times 3 \times 10$ configurations. The 16 pairs are explained in Figure 2 and the $x$-axis expresses their relative rank after ordering the average RMSE over all calibrators; the $3 \times 10$ cross product of AUC and $n$ values are listed on the top and left of the subfigures, respectively. (c)–(d) illustrate RB$^{ind}$ and RB$^{sub}$ with same convention as (a)–(b). Subfigure (e) summarizes (a)–(d) by illustrating the average of: RMSE$^{ind}$ and RMSE$^{sub}$ (left), and RB$^{ind}$ and RB$^{sub}$ (right). The averaging is taken over all configurations conditionally on each value of $n$.

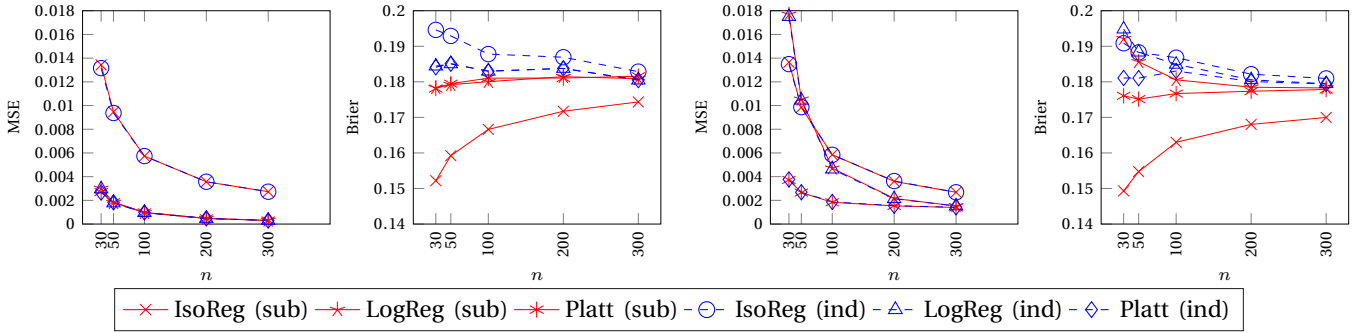

Figure 4: An exact reproduction of the experiments in [4, 5] in terms of MSE and Brier (rather than their root) for three calibrators, on Normal (first two) and Beta (second two) distributions. The two version LogReg and Platt are almost identical.

exhibited from the expanded feature space.

A very interesting observation is that in terms of RMSE, the calibrator's performance gets worse with distribution pairs 11–16. From Figure 2, those are the six pairs involving normal distribution. Not to accuse the normal distribution, the second best distribution pair (and the first best for LogReg) is distribution pair 2, two of whose components are normal (**d, d**). On the other hand, in terms of RB$^{ind}$, the distribution pairs almost do not affect the calibrator's performance. We think that both observations are very counter-intuitive, and no qualitative interpretation is ready off the shelf. Rather, it requires a decomposition to the components of variance under the corresponding parametric distribution pairs, which is quite out of the scope of the present article and is deferred to future work.

It is obvious that RB$^{sub}$ is optimistically biased, which complies with the conventional wisdom in the field of ML for any performance measure. However, this wisdom is sur-prisingly violated when comparing RMSE$^{sub}$ with RMSE$^{ind}$ for almost all calibrators. It seems that RMSE$^{sub}$ is at least the same as, or pessimistically biased than, RMSE$^{ind}$ for most configurations. A similar observation is reported by [4], where they provided their interpretation, to which we agree. Indeed, the Brier score is itself the objective function that is minimized for the different regression methods that will learn from the truth of the class indicator variable. This indicator variable is, as well, the one involved in the Brier score definition (5b). This will cause the usual optimistic/pessimistic bias of the resubstitution/independent testing. In contrast, for the RMSE, the truth of posterior probability (the variable involved in the MSE definition (5a)) is not involved in the training phase. Hence, the estimated RMSE using resubstitution will differ only from that using independent testing by the variance exhibited from the finite sample size of the testing dataset.

*2) PROPROC Calibrator:* The PROPROC software version of [12] calculates the LHR (1b) (Eq. (4a) in [12]), which is

necessary for calibrating new scores after fitting the ROC curve. This software version was used by the authors of [4, 5]. However, this version is neither publicly available at the official website of the University of Chicago nor by personal communication with the developers and administrative staff. After the inventor of the PROPROC, Charles E. Metz, passed away, it seems that the publicly available version of PROPROC is frozen and is not updated. Therefore, the best that we can do is to compare our nine calibrators to the PROPROC results published in [4, 5] using their limited number configurations. This requires reproducing their results so that we can compare all methods on the same workflow. Figure 4 uses their visualization method to illustrate a reproduction of their results, on the two distributions they considered: normal and beta; we used the calibrators LogReg, Platt, and IsoReg. It is clear that the two versions of logistic regression, LogReg and Platt, are almost identical and compare well to the PROPROC for the normal distribution. However, Platt outperforms the others at small $n$ for the beta distribution; then, all converge to the same performance as $n$ increases. Of course, this comparison is very vulnerable to criticism because these calibrators may not compare well on other configurations. However, because of the unavailability of the PROPROC, no more conclusions can be reached.

*3) Multi-Score Calibration: With/Without Expanded Features:* As discussed in Sec. III-B, the only candidate for multi-score calibration is the logistic regression. Therefore, we construct the following set of experiments to study the effect on this calibrator when expanding the feature space; that is we compare LogReg to LogRegExt. Suppose we have two classifiers (scoring functions) $h_1$, $h_2$. Because we have two classes $\omega_0$, $\omega_1$, we then have four distributions: $F_{ij}, i = 0, 1, \ j = 1, 2$, where $F_{ij}$ is the distribution of the scoring function $h_j$ under the class $\omega_i$. We choose each of $F_{ij}, i = 0, 1, \ j = 1, 2$, from the list of the four basic distributions $\{\mathbf{a}, \mathbf{b}, \mathbf{c}, \mathbf{d}\}$ presented at the top of Figure 2. Therefore, we have a total of $4^4$ combinations of distribution pairs. We always adjust the parameters of $F_{0j}, j = 1, 2$, to have 0 mean and unit variance. We adjust the parameters of $F_{1j}, j = 1, 2$, to have unit variance and to achieve a desired separation (expressed in terms of the AUC) with $F_{0j}$. We consider the three different values of AUC used in the single-score calibration. In addition, if two classifiers provide scores for the same problem, we anticipate that there should be some correlation between their scores, because the two classifiers are initially designed to solve the same problem. We account for this by simply assuming a correlation coefficient $\rho$ between the scores $h_j|\omega_0 \sim F_{0j}, j = 1, 2$ (for short, $h_{01}, h_{02}$) and the scores $h_j|\omega_1 \sim F_{1j}, \ j = 1, 2$ (for short, $h_{11}, h_{12}$). We construct this correlation by building the below model. For class $\omega_0$, we consider the two r.v. $v_1 \sim F_{01}$ and $v_2 \sim F_{02}$; then, we construct the scores by the multivariate transformation

$$h_{01} = v_1, \tag{12}$$

$$h_{02} = v_1\rho + v_2\sqrt{(1-\rho^2)}. \tag{13}$$

It is quite straightforward to show that $h_{01}$ and $h_{02}$ have a correlation coefficient of $\rho$. Similarly, we construct $h_{11}$

and $h_{12}$ after adjusting the parameters for the required AUC. We experiment with $\rho = 0$, 0.5, 0.9 to express no, weak, and strong correlations, respectively. To summarize, we have a total of 23,040 experiments as a cross-product of $4^4$ distribution pairs, three values of AUC, three values of $\rho$, and 10 values of sample sizes $n$.

The calibrators are LogReg with the two plain features (scores) $h_1$ and $h_2$ and LogRegExt with the expanded features $h_1$, $h_1^2$, $h_2$, $h_2^2$, $h_1 h_2$ (Sec. III-B2). We only experiment with a second degree polynomial to validate the argument that LogRegExt can work better than LogReg. However, for real problems, practitioners should include several sets of expanded features and keep increasing the complexity to find the optimal space using the typical resampling approaches, for example CV or bootstrap.

The results of these experiments are illustrated in Figure 5. Subfigures (a)–(d) (supplementary material) illustrate the $\text{RMSE}^{ind}$, $\text{RMSE}^{sub}$, $\text{RB}^{ind}$, and $\text{RB}^{sub}$, respectively, of LogRegExt versus LogReg on all configurations. Therefore, each plot displays $4^4 \times 10$ points at particular $\rho$ and AUC. A common heat map of the subfigures codes the sample size $n$, and is displayed under subfigure (e). The latter summarizes subfigures (a)–(d) by plotting the average of each of the four performance measures for each calibrator versus $n$. The average is taken over all the $4^4 \times 3 \times 3$ configurations at a particular sample size $n$.

From the summary subfigure (e), it is obvious that the expanded features enhances the LogRegExt only asymptotically and marginally. However, from the detailed figures (a)–(b), it is obvious that in many configurations, the LogRegExt has a lower error than LogReg (points located below the $y = x$ line.) The majority of these points are red (large $n$) and occur at high $\rho$ and both medium and large AUCs. This is not surprising because at a larger sample size, the regression model can benefit from the expanded feature space without overfitting. At a stronger correlation, there will be more information gained from the expanded feature $h_1 h_2$, which will boost LogRegExt. However, this little improvement may be washed out at smaller AUC values because the problem becomes harder. At this low AUC, the two LH functions are very interleaved; hence, the LHR will be close to 1, which gives a posterior close to 0.5 and average Brier score close to 0.5 as well. This is evident from subfigures (c)–(d). We observe again that both $\text{RMSE}^{ind}$ and $\text{RMSE}^{sub}$ are almost identical. Subfigures (c)–(d) show that in terms of RB, LogReg is better than LogRegExt in almost all configurations and that the expanded features did not boost the latter except at very small fraction of configurations with very little improvements.

*4) Multi- vs. Single-score Calibration:* To test its utility, we compare the multi-score calibration of logistic regression to its single-score calibration. In both cases, we compare its performance with and without feature expansion. We explain how we establish and visualize this comparison for LogReg, and the same applies immediately to LogRegExt. Consider a configuration used to generate a single point on Figure 5a; we denote the performance of LogReg (the $x$ coordinate of this point) by $r_{12}$, and the subscript refers to the fact that the performance is obtained from the multi-score calibration of the two features $h_1$ and $h_2$. Under the aforementioned configuration, we calibrate LogReg using each of the single score $h_1$ or $h_2$ individually; we denote these performances by $r_1, r_2$. Then, for all configurations, we plot $r_{1,2}/r_2$ versus $r_{1,2}/r_1$. We do that for all the eight combinations {RMSE, RB} × {LogReg, LogRegExt} × {ind, sub}.

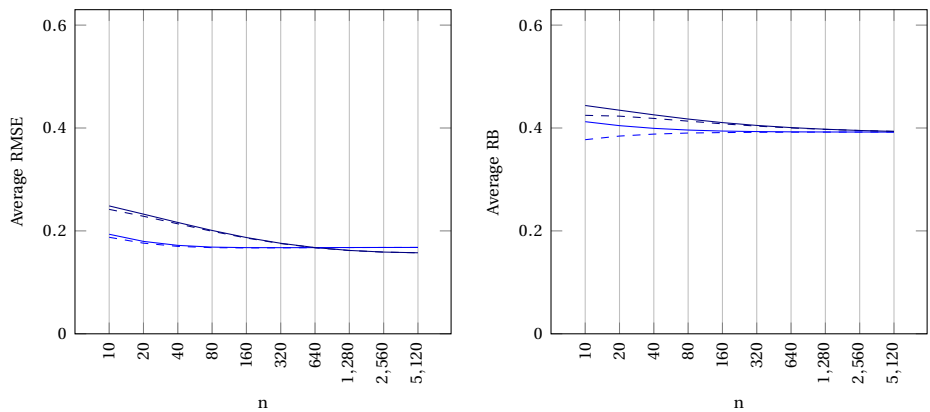

Figure 5e: multi-score calibration: summary plot for (a)–(d).

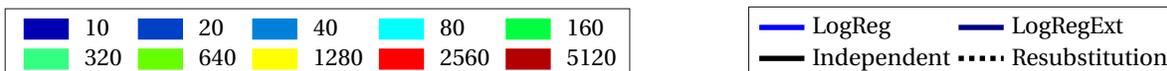

Figure 5: consists of 5 subfigures (a)–(e) with two legends: (left) the heat map that codes the sample size $n$ for subfigures (a)–(d) and (right) calibrators for subfigure (e). (a)–(d) illustrate the RMSE$^{ind}$, RMSE$^{sub}$, RB$^{ind}$, and RB$^{sub}$, respectively, of LogRegExt vs. LogReg for all configurations. Each plot consists of 3 × 3 plots; each plot illustrates $4^4 \times 10$ configurations (points) at particular $\rho$ and AUC. Subfigure (e) summarizes (a)–(d) by illustrating the average of: RMSE$^{ind}$ and RMSE$^{sub}$ (left), and RB$^{ind}$ and RB$^{sub}$ (right). The averaging is taken over all configurations conditionally on each value of $n$.

The results of these comparisons are illustrated in Figure 6, which consists of nine subfigures that follow the same plotting style, convention, and legends of Figure 5. The first eight subfigures (supplementary material) correspond to the eight combinations, respectively, with the order of the flattened cross-product. If the multi-score calibration is better than each individual single-score calibration, that is $r_{1,2} < \min(r_1, r_2)$, then the point will be inside the unit square. The mark + inside the unit square is located at position $(p, p)$, where $p$ is the fraction of points located inside the unit square. Therefore, $p$ represents how frequent the multi-score calibration wins over both single score calibrations. The summary of the eight subfigures (a)–(h) is illustrated in subfigure (i), which plots the average $p$ versus $n$ for each of the eight combinations, and the average is taken over all configurations of each subfigure conditionally on each value of $n$.

From the summary subfigure (i), in terms of the RMSE, the multi-score calibration improves LogReg only in roughly less than 7% of experiments and almost constantly with the sample size $n$; it improves LogRegExt in less than 9% of experiments at $n = 320$. In terms of RB, the multi-score calibration improves the performance of at least 80% of the experiments with a small $n$ and almost 100% of experiments with $n > 40$. The detailed subfigures (a)–(h) reveal more. (a) and (c) shows that for LogReg and LogRegExt, respectively,

$p$ is small at low AUCs, regardless of $\rho$. As explained above, when we commented on Figure 5, the calibration problem itself is hard, and the posterior is near 0.5. This will leave a small margin of improvement. However, at higher AUCs and a low correlation coefficient $\rho$, the fraction $p$ increases for LogReg and increases significantly for LogRegExt. This increase of the value of $p$ retracts again when increasing $\rho$. The interpretation is that at a high AUC, the problem is easier for each single-score calibration, and at a high $\rho$ the two scores will almost behave the same; hence, combining both scores in the multi-score calibration may not be of great value. (e)-(h) illustrate the RB and show that, almost for the majority of the experiments, $p \simeq 1$. The improvement factor is significant, almost at 0.75, where the majority of points is located at the upper right corner of the unit square.

### B. Real Datasets

We carried out experiments, similar to those carried out in the previous section, on three real datasets from the domain of cybersecurity: TREC-05 dataset [30], [31], Drebin dataset [32], [33], and the SSDP Flood dataset used in [34]. We interpret the results here and defer further investigations and interpretations to the discussion in Sec. V.

*1) TREC-05 Dataset:* TREC-05 is a popular public email security dataset consisting of 72,450 emails labeled as spam or not spam. Because the dataset is very large, we have the luxury of splitting it into training and testing to construct the classifiers before the calibration process. Therefore, we split the dataset into two halves; each has 36,225 samples. Support vector machine (SVM) and random forest (RF) classifiers were trained on $n$-gram features extracted from the training dataset; then, both classifiers were frozen, and the training set was never used in the rest of experiments. The SVM and RF were tested on the testing set, and each produced 36,225 scores. These scores have a correlation coefficient of 0.91, and each achieved, respectively, 96.6% and 97.53% balanced accuracy, and 0.994 and 0.996 AUCs. The smoothed histograms of these scores, under the two classes, are shown in Figure 7. For the RF, these histograms are abruptly exponential-like and surprisingly very different from the simulated datasets in the previous sections. It is worth mentioning that [35] proved that the classifier scores can be abruptly exponential even when the training features are normally distributed.

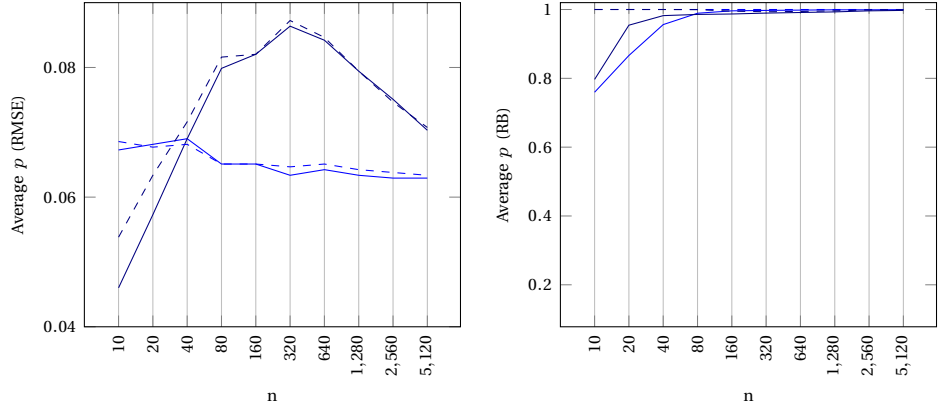

Figure [6]i: multi- vs. single-score calibration: summary plot for (a)–(h).

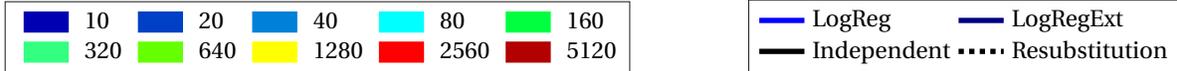

Figure 6: consists of 9 subfigures, with the same plotting style, convention, and legends, of Figure [5]. The first 8 subfigures correspond to the 8 combinations of the cross product {RMSE, RB} × {LogReg, LogRegExt} × {ind, sub} as indicated by the sub-caption. Each subfigure consists of $3 \times 3$ plots (for $\rho$ and AUC combinations); each plot illustrates $4^4 \times 10$ point (for distribution pairs and $n$ combinations); each point is $r_{1,2}/r_1, r_{1,2}/r_2$, were $r_1, r_2, r_{1,2}$ are, respectively, the performance of the calibrator using each of the single-score and the multi-score. The symbol + inside the unit square is located at $(p, p)$, where $p$ is the ratio of the points insde the unit square to the total points.

It is interesting to study the behavior of different calibrators on different sample sizes obtained from this dataset. Said differently, we will pretend we are in a situation where we have obtained only $2n$ observations from this dataset. Then, to train the calibrators on different training sets of the same size $2n$, to mimic a MC trial and to further test the trained calibrator on a large independent testing set, we proceeded as follows: Out of the 36,225 scores available from testing each classifier, we sequestered half to account for the independent testing sets for calibrators. Then, for each value of $n = 10 \times 2^i$, $i = 0, \cdots, 9$, we randomly selected $M(=1000)$ training sets, each is of size $2n$ with $n$ malicious and $n$ benign samples. Then, we trained the calibrators on these training sets. We only report RB$^{ind}$ and RB$^{sub}$ because RMSE is not available because of the absence of the true posterior probability. We use the same calibrators and the same comparative study used for the simulated datasets above.

The results of these experiments are illustrated in Figure 8, which plots RB$^{ind}$ and RB$^{sub}$ versus the sample size $n$ for three cases: single-score calibration of each of SVM and RF scores, and multi-score calibration using both. The results are different from those obtained from the simulated datasets. For the single-score calibration, although all calibrators converge almost to the same performance as $n$ increases, at a small $n$ the simple BinReg competes with IsoReg, and the third place is LogReg or Platt for SVM or RF, respectively. Platt almost wins over the rest of the values of $n$. LogReg and LogRegExt are the worst two. However, LogRegExt is better than LogReg for only RF scores. The multi-score calibration with either LogReg or LogRegExt wins over all the single calibrators except at small values of $n$. This once more shows the value of our proposed method of multi-score calibration in many (not all) cases.

*2) Drebin Dataset:* Drebin is a popular malware analysis dataset consisting of features extracted from 129,013 android applications, 5,560 of which are malicious applications belonging to 179 different malware families, while the rest are benign. From each application, 545,334 binary features are extracted, indicating the presence of various API calls, permissions, network addresses, and so forth. Because of the size of the dataset, it is not available in one large file. Instead, for each sample, the plain text representation of each present feature is listed in a text file; features that are not present are not listed. This allows for a very condensed representation of the dataset because most samples contained well under 100 features.

To establish some basis of comparison, we used SVM and RF for this dataset. Because these algorithms did not support online (incremental) learning, the dataset had to be manually extracted into one large matrix using the sparse matrix representation of Scipy. If the data set was defined in a standard numpy array, then it would require about at least 70 GB just to fit in the memory (assuming a byte-sized datatype for each element).

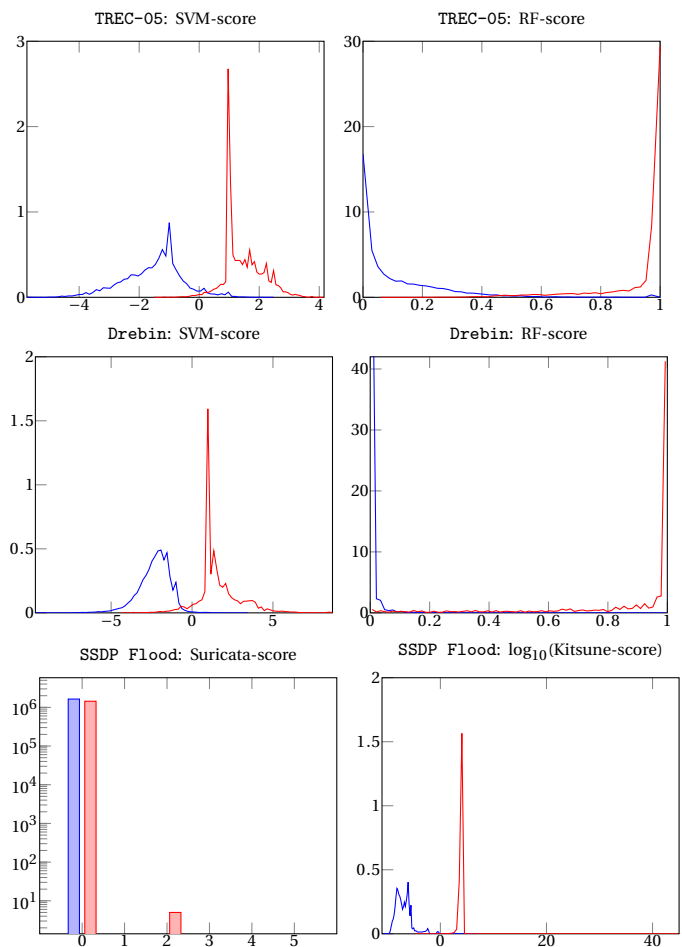

Figure 7: Smoothed histogram of the scores of the two classes $\omega_0$ (blue) and $\omega_1$ (red) of the three datasets: SVM and RF of TREC-05 and Drebin (row 1 and row 2), and Suricata and Kitsune of SSDP (row 3). The AUC of the six plots are: 0.994, 0.996, 0.992, 0.996, 0.5, 0.999, respectively. For each dataset, the correlation coefficient between its two sets of scores is 0.91 and 0.64, and $2 \times 10^{-6}$, respectively.

We used the `Scikit` implementation (with `liblinear` for the SVM) of each classifier and trained them using 10-fold CV, while the predicted scores on the left-out folds were used for calibration. This is because the sample size of the malicious class is only 5,560. We adjusted the classification threshold so that the FPF was 1 in 100 (the same FPF chosen in [32]). Our SVM and RF achieved a 0.955 and 0.971 detection rate, respectively. This is higher than then the reported detection rate of 0.939 in [32], where they used the SVM. The scores of the SVM and RF achieved an AUC of 0.992 and 0.996, respectively, and their correlation coefficient is 0.64. Figure 7 illustrates the histogram of the scores; they almost show the same pattern as those of the TREC-05 dataset.

The results of these calibration experiments are illustrated in Figure 8 and are plotted in the same way as those of TREC-05 dataset. The results are very similar but have some differences. For single-score calibration, all the methods, except LogRegExt on SVM and both LogRegExt and LogReg on RF, are almost comparable at all values of $n$. If compared with their performances on the RF scores of TREC-05 dataset, it seems that LogReg and LogRegExt switched places. However, LogReg on the multi-score calibration is almost the overall winner.

*3) SSDP Flood Dataset:* was obtained by monitoring network traffic while executing a Denial of Service attack in an IP camera video surveillance network. This type of attack overloads the digital video recorder by causing cameras to spam the server with Universal Plug and Play messages. The dataset consists of a single `.pcap` file which contains 4,077,266 packets, and a label vector, indicating whether each packet is benign or malicious. On this dataset we used Suricata [36] and Kitsune [34], two popular network intrusion detection systems (NIDS), to provide two independent scores for each observation. Suricata is signature based, which uses predefined handcrafted rules to determine if a packet is a threat. It generates an alert and assigns the packet an integer threat score in the range 1–5, with higher scores representing a more serious threat. We used the most recent version of Suricata [36] and of the Emerging Threats Open rule set [37]. Kitsune, in contrast to Suricata, is anomaly based and gives a continuous score in the range $[0, \infty)$, with higher scores corresponding to more anomalous and presumably more likely malicious traffic. Because Kitsune's scores lie in such a wdie range, we use the log of the score for computational stability. Signature based NIDS typically have a much lower false positive rate than anomaly based NIDS; however, the former are incapable of detecting novel threats in contrast to the latter. Thus we selected two different detection systems, which when used together can hopefully account for the weakness of each other. Both systems are online, and therefore can detect threats in real time. Thus we think this experiment represents a practical real world use-case for the calibration of multiple classifiers.

Following the training method used in the original Kitsune paper, we used the first 1,000,000 packets in the `.pcap` file (assumed to be all benign traffic) to train Kitsune. Threat scores from Suricata and Kitsune were then obtained for the remaining 3,077,266 packets, 1,439,604(46.8%) of which represent malicious traffic. Figure 7 shows the bar chart of the discrete Suricata scores (the two classes are very overlapped in the category 0) and the smoothed histogram of the Kitsune scores (the two classes are very discriminated), with an AUC of 0.500002 and 0.99999, respectively. The very low AUC of Suricata means that its decision on this dataset is almost a random guess, which explains why the scores of these two systems are completely uncorrelated with a correlation coefficient of $2 \times 10^{-6} \simeq 0$.

We applied to this dataset the same method that we applied to TREC-05 and Drebin above for evaluating the different calibration techniques. However, since we have such a large test set, we used a histogram to derive the "pseudo true posterior" from this test set and used RMSE, in addition to RB, to compare the predicted posteriors obtained from calibration with the pseudo true posterior. The results of these calibration experiments are illustrated in Figure 8.

For single-score calibration with Suricata's scores, because they were nearly random, all calibrators correctly learn to predict a 50% probability of maliciousness for each sample ($\hat{p} \simeq 0.5$), even at $n = 10$. Thus all methods achieve RB $\simeq 0.5$ and RMSE $\simeq 0$ across all experiments. This clearly indicates the importance of calibration in the cybersecurity setting. A practitioner seeing a threat score of 0 out of 5 may erroneously assume it a negligible threat, despite of a 50% chance of malicious traffic; but after seeing the calibrated score the practitioner would learn not to rely solely on Suricata before concluding that their systems are safe.

For single-score calibration with Kitsune's scores, all methods correctly learn the pseudo-true posterior asymptotically with $n$. Once again, we see the usefulness of score calibration in the cybersecurity setting. With the uncalibrated score, the security practitioners have no way of assessing Kitsune's output without comparing with previous outputs, because the output is on an arbitrary scale; all they know is that the higher the number the bigger the threat. With score calibration, the practitioner only needs to check one number conveniently representing the probability of a true threat.

In the multi-score calibration experiments, it is remarkable that the calibrators correctly ignore the noninformative Suricata scores and achieves a stable performance that is identical to that of Kitsune's single-score calibration experiments.

### C. Computational Power and Execution Complexity

All experiments on the simulated datasets were carried out on the Compute Canada Cedar Cluster[3]. For the single-score calibration (Sec. IV-A1), the experiments were run using a job array for scheduling multiple independent jobs (scripts), with all jobs running in parallel. We wrote 48 separate jobs, one job for each configuration of the cross-product of the 16 distribution pairs and three AUC values. Each job accounts for the 10 values of $n$. The job took about an hour on one core, 8 GB of RAM, obtaining 96% CPU utilization, and 63% memory utilization. Only the single-score simulation for Platt's sigmoid fit was carried out separately on an eight-core machine with 16 GBs of memory and took 36 hours to complete using six threads, with a thread per job.

For the multi-score calibration (Sec. IV-A3), a job array with 768 jobs was created, one job for each configuration of the cross-product of the $4^4$ distribution pairs and three AUC values. Each job accounts for the cross-product of three values of $\rho$ and 10 values of $n$. The whole job array took five hours to execute, although the majority of jobs were executed in three hours. On average, these jobs obtained 94% CPU efficiency and 62% memory efficiency with one core and 8 GB of RAM per job. These experiments obviously took longer than their counterparts for the single-score calibration for the relative number of configurations (23,040 vs. 480).

## V. Discussion

The empirical distributions in Figure 7, for the real datasets, show two main differences from the simulated data in the previous section: (1) it is exponential-like for the RF scores and (2) the AUC is very high, 0.994, 0.996, 0.992, 0.996, respectively, for the combinations {TREC-05, Drebin} × {SVM, RF} appearing in the figure). Therefore, we designed the following simulation experiment for a deeper investigation. Let $F_0$ be the truncated exponential distribution and $F_1$ be a flipped truncated exponential distribution, both share the same population parameter $\lambda$ and the finite support $[0, 1]$ so that they look symmetric. The population parameter $\lambda$ is adjusted to obtain a required AUC, which takes the same three values of the previous simulated data, in addition to the value 0.99.

The results of this experiment are illustrated in Figure 9. The RMSE$^{ind}$, RMSE$^{sub}$, RB$^{ind}$, and RB$^{sub}$ of the nine calibrators are plotted versus the sample size $n$ at different AUC values. One more surprise is that the performance for the first three values of AUC complies with the simulated data in the previous section with some performance deterioration observed for LogReg. Only the performance at the very high AUC of 0.99 looks similar to the performance of the calibrators on the real dataset. Still, comparing the results on Figure 9, but here recalling the results of the real datasets on Figure 8 at this very high AUC, both LogReg and LogRegExt are inferior to other methods, including the Platt version of logistic regression, exactly as was observed for the real dataset. At this very high AUC, the two distributions are very separated, which is an ideal situation to exhibit the value of the bias correction term of Platt's version of logistic regression. IsoReg always has a moderate performance among others.

It is quite important to note to that obtaining such a very high discriminating power, represented as AUC = 0.99+, is very unrealistic in the vast majority of ML applications. The real datasets above, where both SVM and RF were able to provide a very high AUC, look very clean and atypical. Such a very high AUC means that the two sets of classification scores are almost perfectly separable such that any threshold will provide near 100% TP and 0% FP.

We try to summarize the findings and the final message of Figures 3–9 to finally provide the advice for practitioners. It is almost clear that there is no over all winner across all possible configurations. However, major remarks are in order.

- The Platt version of the logistic regression shows superiority over the default version that exists in the scientific libraries, for example `scikit`, at some special configurations. Therefore, it is worth implementing it in an optimized way in these libraries to replace the default version; and it is better to leverage it in the calibration process.
- Among the single-score calibrators on the plain features, that is without feature expansion, the Platt method is the winner in the vast majority of configurations, even if it is not the over all winner. It only looses in a few experiments with little margin.
- The isotonic regression, although it rarely wins in contrast to Platt, it rarely deteriorates similarly to Platt.
- The PROPROC method does not seem to be feasible because the software version that can calibrate scores

is no longer publicly available.

- The very rare cases with a very high AUC should not be taken as a basis for comparing the calibrators. These are very special cases, and even through we found them in some real datasets, they should be dealt with on a case-by-case basis. For example, the practitioner should run a Mann-Whitney test of the scores to estimate the AUC and then decide whether it is exceptional or not.
- The multi-score calibration boosted the performance of LogReg for some experiments in terms of RMSE and almost for all experiments, including the real dataset in terms of RB.
- The feature expansion boosted the performance of a small fraction of experiments in terms of RMSE and a negligible faction of experiment in terms of RB.

Based on all of these observations, if we have to provide only a single piece of advice to practitioners, we would recommend the Platt version of the logistic regression for a single-score calibration without feature expansion. When there are more classifiers providing the score for the same problem, we would recommend the Platt regression using a multi-score calibration without feature expansion as well.

However, for a more mature practitioner, we need to recall the observation that there is not an overall winner among calibrators and configurations. Therefore, we may include more than one calibrator, including Platt, on the expanded features and incorporate all of them in a CV model selection per the usual paradigm in training and testing ML models.

## VI. Conclusion

The current paper studied the calibration of binary classifiers, where the classifier score of a testing observation, does not necessarily correspond to its posterior probability of belonging to one of the two classes. The paper surveyed the literature, summarized the different methods available for calibrating a single classifier, and formalized the mathematical properties of the calibration process, for example the invariance of some classification performance measures. The paper presented three main contributions: (1) proposing the multi-score calibration, that is designing a calibrator when more than one classifier provides the scores to a single observation; (2) treating the score(s) as features and expanding them to higher dimensions to design a calibrator; (3) conducting a massive number of experiments on simulated datasets, in the order of 24,000 experiments, and on two real datasets from the cybersecurity domain to study the behavior of different calibrators at different configurations. A final piece of advice for practitioners is provided based on the visualization and interpretation of the results that show there is no overall winner among the different calibrators. However, in general, (1) the Platt's version of the logistic regression has a decent accepted performance over a wide range of experiments; (2) the multi-score calibration boosts the performance in the majority of experiments; and (3) expanding the score to higher dimensions helps in some experiments.

## VII. Acknowledgment

We would like to thank our friends Dr. Weijie Chen, of the US FDA, and Dr. Lorenzo Pesce, formerly the developer of PROPROC software at University of Chicago, for their long discussions on the software. Special thanks to Dr. Weijie Chen for sharing with us his Matlab code of his publications [4, 5].

## References

[1] J. Platt *et al.*, "Probabilistic outputs for support vector machines and comparisons to regularized likelihood methods," *Advances in large margin classifiers*, vol. 10, no. 3, pp. 61–74, 1999.

[2] G. W. Brier *et al.*, "Verification of forecasts expressed in terms of probability," *Monthly weather review*, vol. 78, no. 1, pp. 1–3, 1950.

[3] A. Niculescu-Mizil and R. Caruana, "Predicting good probabilities with supervised learning," in *Proceedings of the 22nd international conference on Machine learning.* ACM, 2005, pp. 625–632.

[4] W. Chen, B. Sahiner, F. Samuelson, A. Pezeshk, and N. Petrick, "Calibration of medical diagnostic classifier scores to the probability of disease," *Statistical methods in medical research*, vol. 27, no. 5, pp. 1394–1409, 2018.

[5] ——, "Investigation of methods for calibration of classifier scores to probability of disease," in *Medical Imaging 2015: Image Perception, Observer Performance, and Technology Assessment*, vol. 9416. International Society for Optics and Photonics, 2015, p. 94161E.

[6] R. H. Randles and D. A. Wolfe, *Introduction to the theory of nonparametric statistics.* New York: Wiley, 1979.

[7] J. Hájek, Z. Šidák, and P. K. Sen, *Theory of rank tests*, 2nd ed. San Diego, Calif.: Academic Press, 1999.

[8] W. A. Yousef, R. F. Wagner, and M. H. Loew, "Assessing Classifiers From Two Independent Data Sets Using ROC Analysis: a Nonparametric Approach," *Pattern Analysis and Machine Intelligence, IEEE Transactions on*, vol. 28, no. 11, pp. 1809–1817, 2006.

[9] W. Chen, B. D. Gallas, and W. A. Yousef, "Classifier Variability: Accounting for Training and testing," *Pattern Recognition*, vol. 45, no. 7, pp. 2661–2671, 2012.

[10] G. Z. Grudic, "Predicting the probability of correct classification," *Unpublished technical report*, 2004.

[11] B. Zadrozny and C. Elkan, "Learning and making decisions when costs and probabilities are both unknown," in *Proceedings of the seventh ACM SIGKDD international conference on Knowledge discovery and data mining.* ACM, 2001, pp. 204–213.

[12] L. L. Pesce, K. Horsch, K. Drukker, and C. E. Metz, "Semiparametric estimation of the relationship between roc operating points and the test-result scale: Application to the proper binormal model," *Academic radiology*, vol. 18, no. 12, pp. 1537–1548, 2011.

[13] D. D. Dorfman, K. S. Berbaum, and C. E. Metz, "Receiver Operating Characteristic Rating Analysis - Generalization To the Population of Readers and Patients With the Jackknife Method," *Investigative Radiology*, vol. 27, no. 9, pp. 723–731, 1992.

[14] D. D. Dorfman, K. S. Berbaum, C. E. Metz, R. V. Lenth, J. A. Hanley, and H. Abu Dagga, "Proper Receiver Operating Characteristic Analysis: the Bigamma model," *Acad Radiol*, vol. 4, no. 2, pp. 138–149, 1997.

[15] C. E. Metz, B. A. Herman, and J.-H. Shen, "Maximum Likelihood Estimation of Receiver Operating Characteristic (ROC) Curves From Continuously-Distributed Data," *Statistics In Medicine*, vol. 17, no. 9, pp. 1033–1053, 1998.

[16] C. E. Metz and X. Pan, ""Proper" Binormal ROC Curves: Theory and Maximum-Likelihood Estimation." *Journal of Mathematical Psychology*, vol. 1, pp. 1–33, 1999.

[17] X. Pan and C. E. Metz, "The "proper" Binormal Model: Parametric Receiver Operating Characteristic Curve Estimation With Degenerate data," *Academic Radiology*, vol. 4, no. 5, p. 380, 1997.

[18] L. L. Pesce and E. C. Metz, "Reliable and Computationally Efficient Maximum-Likelihood Estimation of "proper" Binormal Roc Curves," *Academic Radiology*, vol. 14, no. 7, p. 814, 2007.

[19] C. A. Roe and C. E. Metz, "Dorfman-Berbaum-Metz Method for Statistical Analysis of Multireader, Multimodality Receiver Operating Characteristic Data: Validation With Computer simulation," *Academic Radiology*, vol. 4, no. 4, pp. 298–303, 1997.

[20] B. Zadrozny and C. Elkan, "Transforming classifier scores into accurate multiclass probability estimates," in *Proceedings of the eighth ACM SIGKDD international conference on Knowledge discovery and data mining*. ACM, 2002, pp. 694–699.

[21] D. Ślezak, A. Chadzyńska-Krasowska, J. Holland, P. Synak, R. Glick, and M. Perkowski, "Scalable cyber-security analytics with a new summary-based approximate query engine," in *2017 IEEE International Conference on Big Data (Big Data)*. IEEE, 2017, pp. 1840–1849.

[22] T. Hastie, R. Tibshirani, and J. H. Friedman, *The elements of statistical learning: data mining, inference, and prediction*, 2nd ed. New York: Springer, 2009.

[23] J. Kittler, M. Hatef, R. P. W. Duin, and J. Matas, "On Combining classifiers," *Pattern Analysis and Machine Intelligence, IEEE Transactions on*, vol. 20, no. 3, pp. 226–239, 1998.

[24] L. I. Kuncheva, "A theoretical study on six classifier fusion strategies," *IEEE Transactions on Pattern Analysis and Machine Intelligence*, vol. 24, no. 2, pp. 281–286, Feb 2002.

[25] W. Li, J. Hou, and L. Yin, "A classifier fusion method based on classifier accuracy," in *2014 International Conference on Mechatronics and Control (ICMC)*, July 2014, pp. 2119–2122.

[26] J. S. Ramberg, P. R. Tadikamalla, E. J. Dudewicz, and E. F. Mykytka, "A Probability Distribution and Its Uses in Fitting Data," *Technometrics*, vol. 21, no. 2, p. 201, 1979.

[27] W. A. Yousef, S. Kundu, and R. F. Wagner, "Nonparametric estimation of the threshold at an operating point on the roc curve," *Computational Statistics & Data Analysis*, vol. 53, no. 12, pp. 4370–4383, 2009.

[28] W. A. Yousef, "Assessing classifiers in terms of the partial area under the roc curve," *Computational Statistics & Data Analysis*, vol. 64, no. 0, pp. 51–70, 2013.

[29] F. Pedregosa, G. Varoquaux, A. Gramfort, V. Michel, B. Thirion, O. Grisel, M. Blondel, P. Prettenhofer, R. Weiss, V. Dubourg, J. Vanderplas, A. Passos, D. Cournapeau, M. Brucher, M. Perrot, and E. Duchesnay, "Scikit-learn: Machine learning in Python," *Journal of Machine Learning Research*, vol. 12, pp. 2825–2830, 2011.

[30] G. V. Cormack and T. R. Lynam, "Online supervised spam filter evaluation," *ACM Transactions on Information Systems (TOIS)*, vol. 25, no. 3, pp. 11–es, 2007.

[31] ——, "Trec 2005 spam track overview." in *TREC*, 2005, pp. 500–274.

[32] D. Arp, M. Spreitzenbarth, M. Hubner, H. Gascon, K. Rieck, and C. Siemens, "Drebin: Effective and explainable detection of android malware in your pocket." in *NDSS*, vol. 14, 2014, pp. 23–26.

[33] M. Spreitzenbarth, F. Freiling, F. Echtler, T. Schreck, and J. Hoffmann, "Mobile-sandbox: having a deeper look into android applications," in *Proceedings of the 28th Annual ACM Symposium on Applied Computing*, 2013, pp. 1808–1815.

[34] Y. Mirsky, T. Doitshman, Y. Elovici, and A. Shabtai, "Kitsune: an ensemble of autoencoders for online network intrusion detection," *arXiv preprint arXiv:1802.09089*, 2018.

[35] W. A. Yousef, "Prudence when assuming normality: an advice for machine learning practitioners," *arXiv preprint arXiv:1907.12852*, 2019.

[36] Suricata 6.0.4. https://suricata.io/download/.

[37] Suricata emerging threat open ruleset. https://rules.emergingthreats.net/OPEN_download_instructions.html.

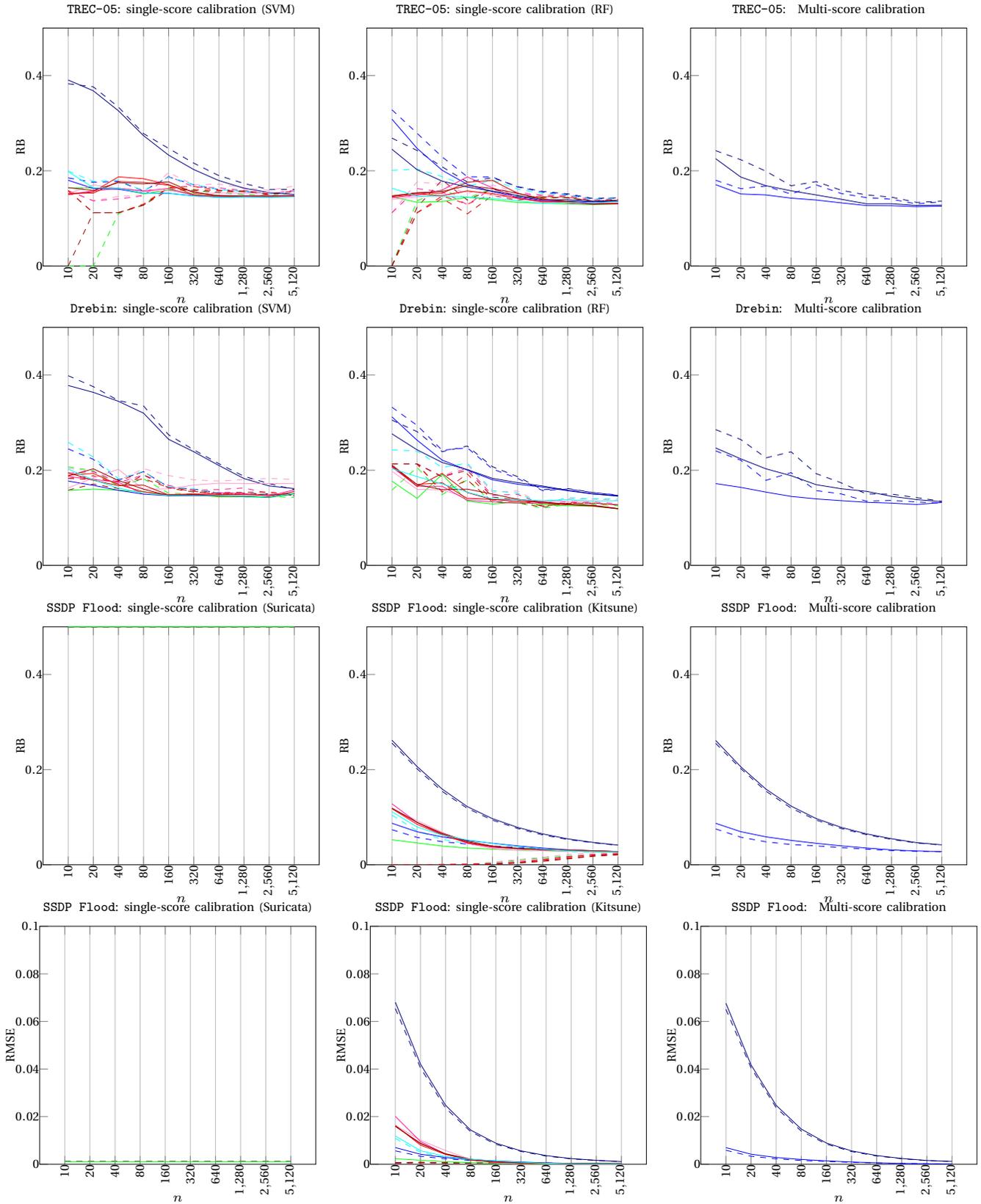

Figure 8: The $RB^{ind}$ and $RB^{sub}$ of all calibrators on the three real datasets TREC-05, Drebin, SSDP (first three rows, respectively), and the $RMSE^{ind}$ and $RMSE^{sub}$ for SSDP (fourth row). On each row, there are three plots corresponding to calibration of the two individual scoring methods (first two columns) and the multi-score of both (third columns).

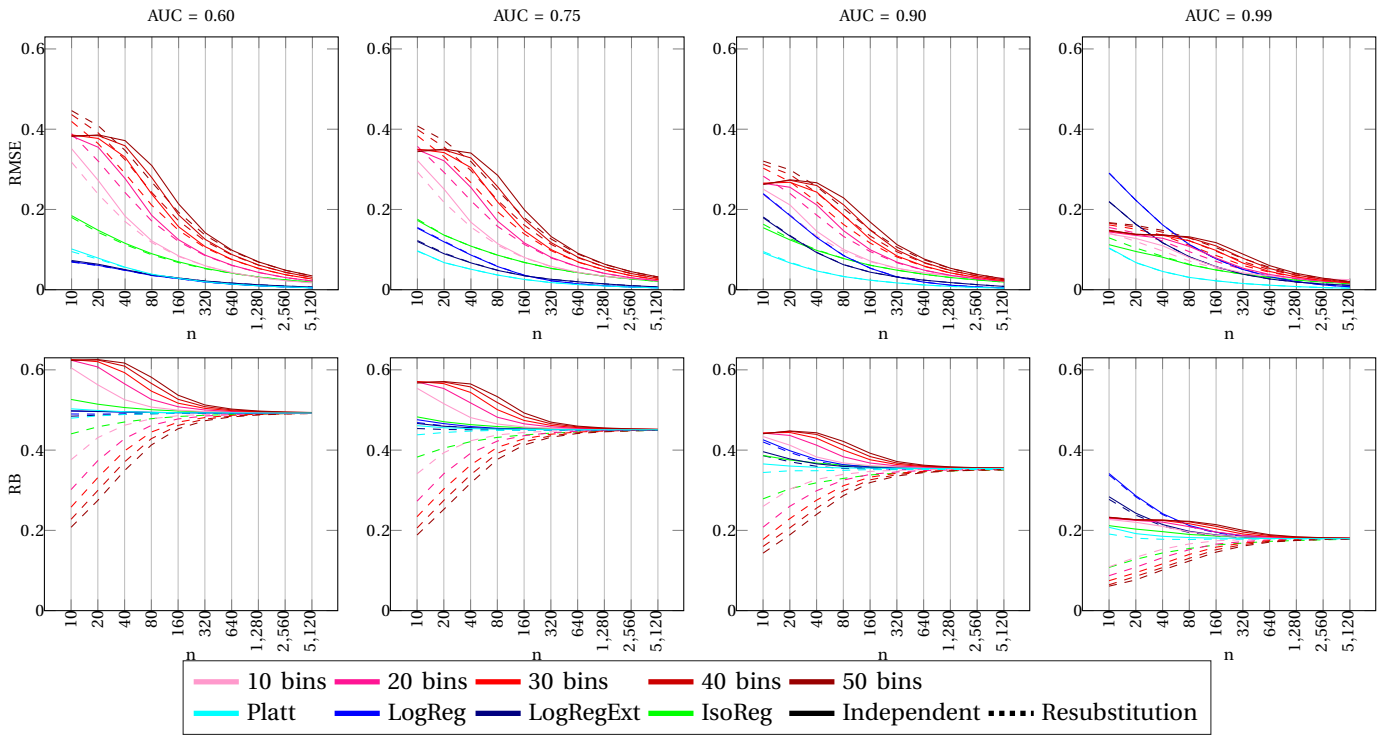

Figure 9: The relative performance of the 9 calibrators using scores simulated from truncated exponential distributions similar to those of RF in Figure 7. The RMSE and RB are illustrated on top and bottom rows respectively. At higher AUC, the superior performance of the LogReg disappears and the Platt version, along with the BinReg are the winners.

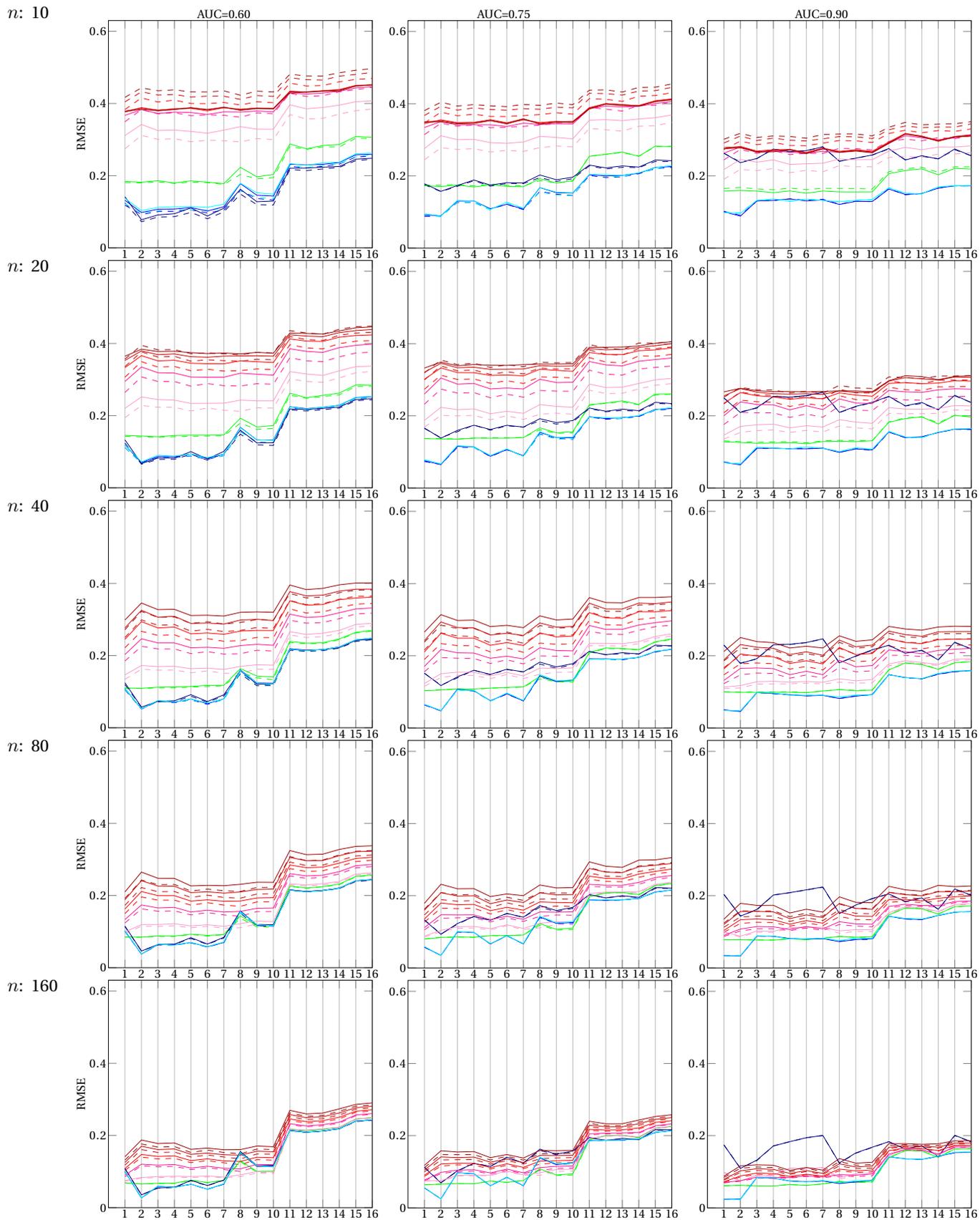

Figure 3a: single-score calibration: RMSE.

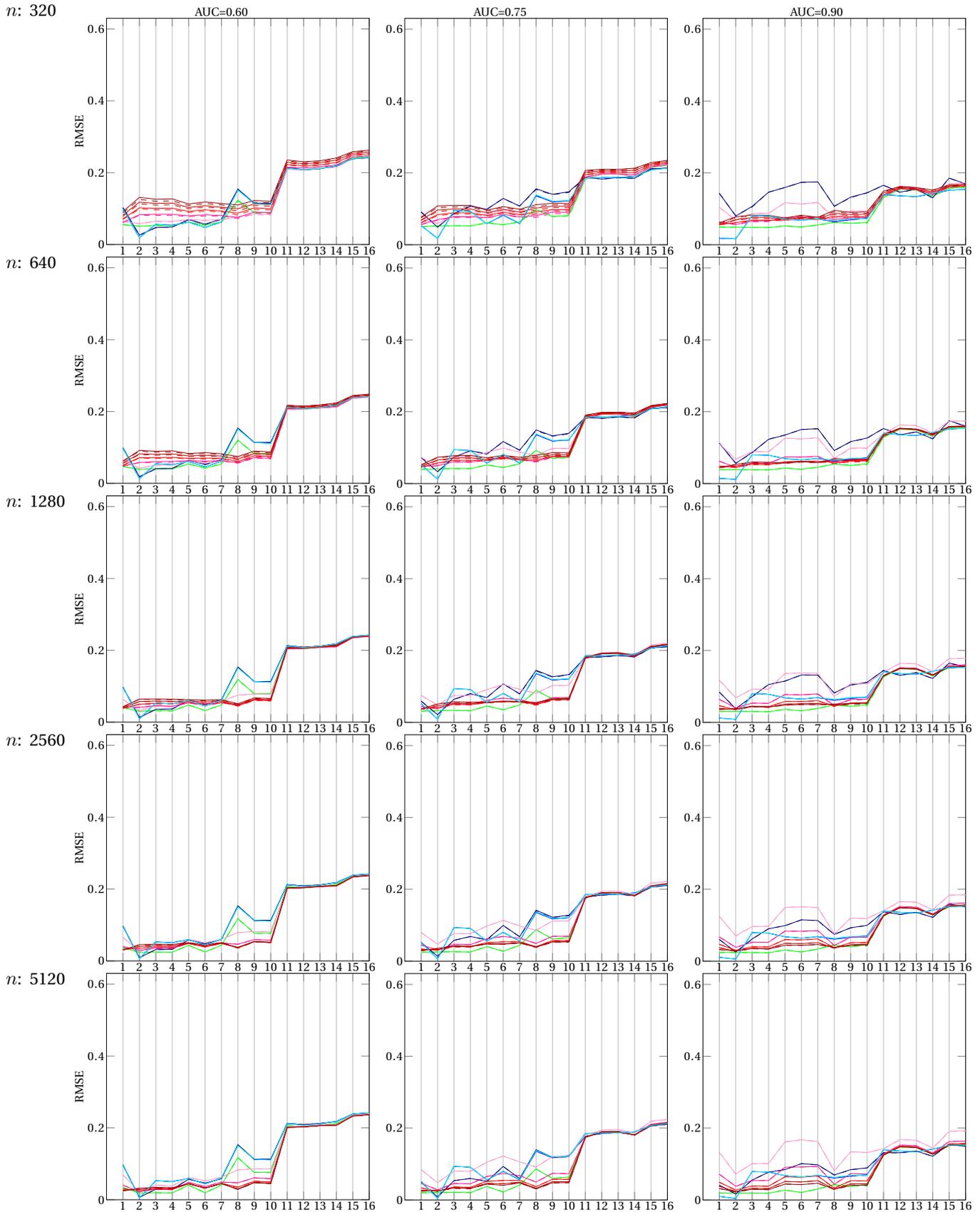

Figure 3b: single-score calibration: RMSE, cont.

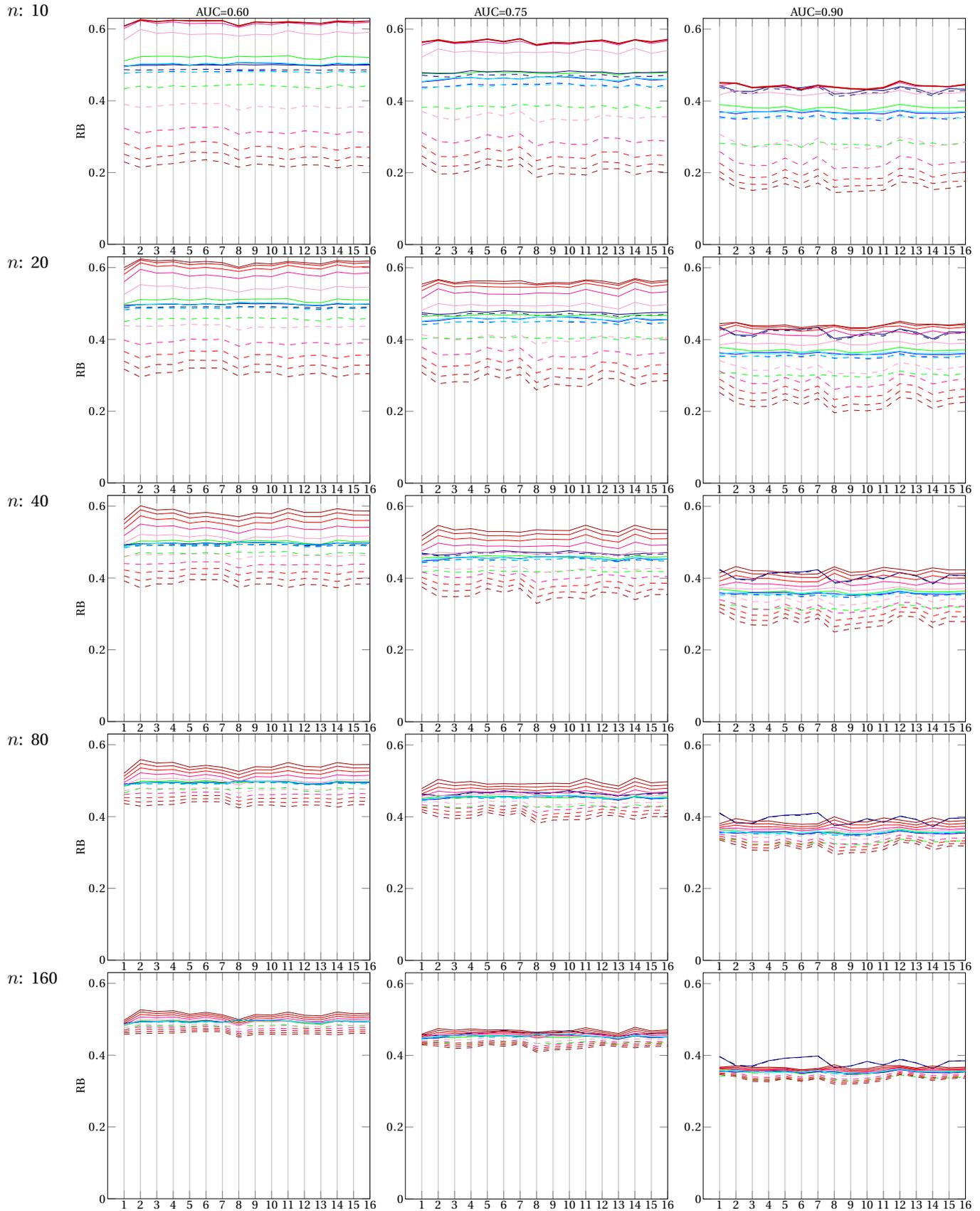

Figure 3c: single-score calibration: RB.

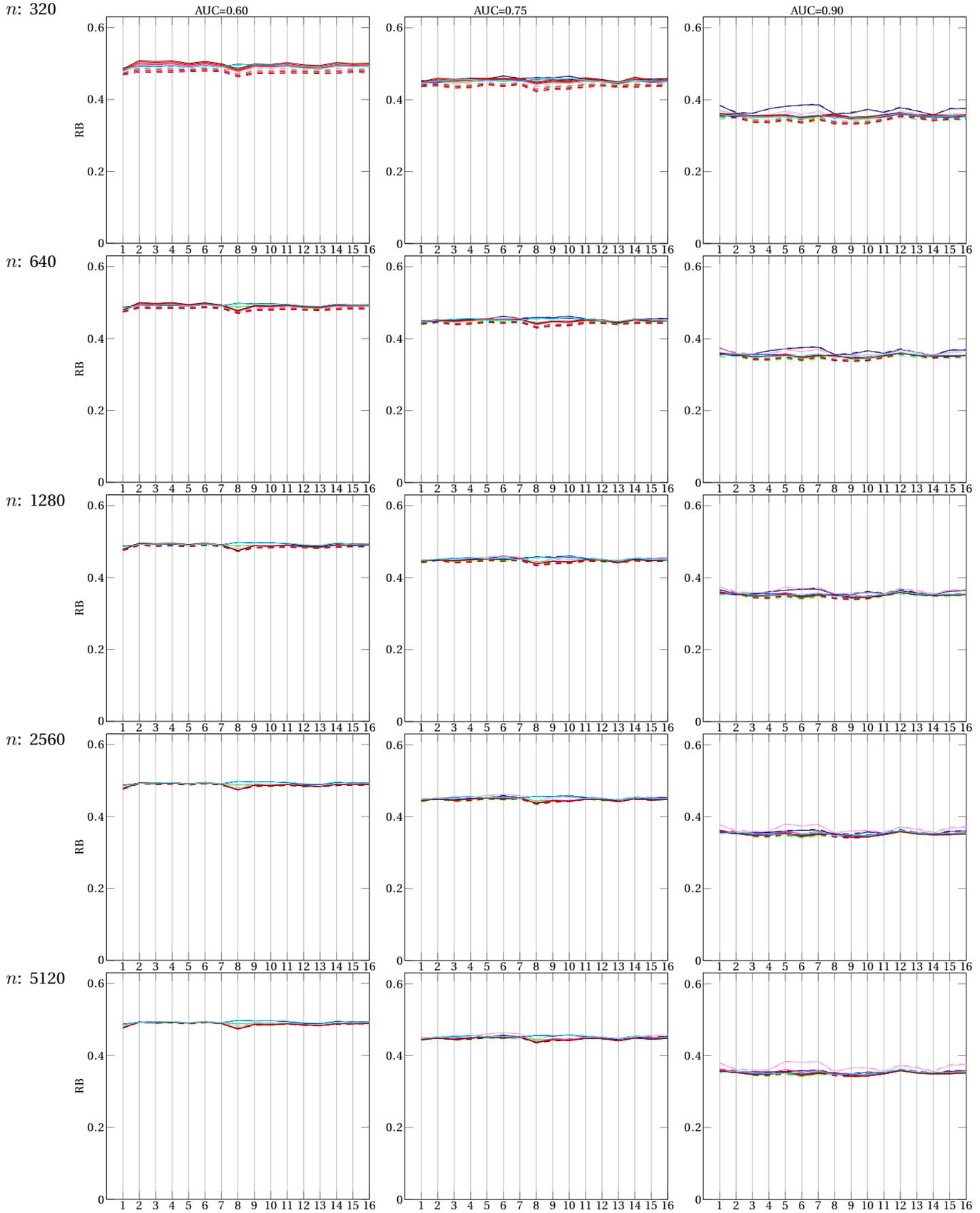

Figure 3d: single-score calibration: RB, cont.

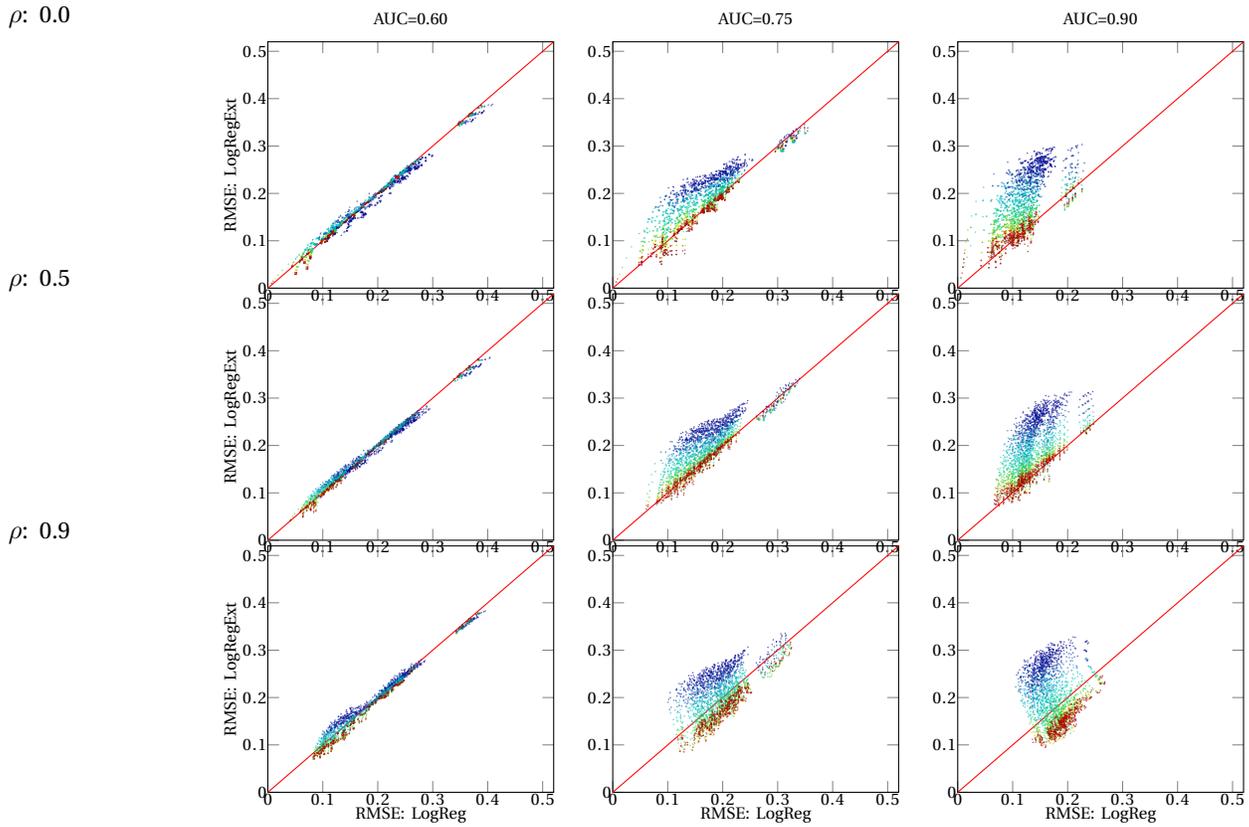

Figure 5a: multi-score calibration: RMSE$^{ind}$.

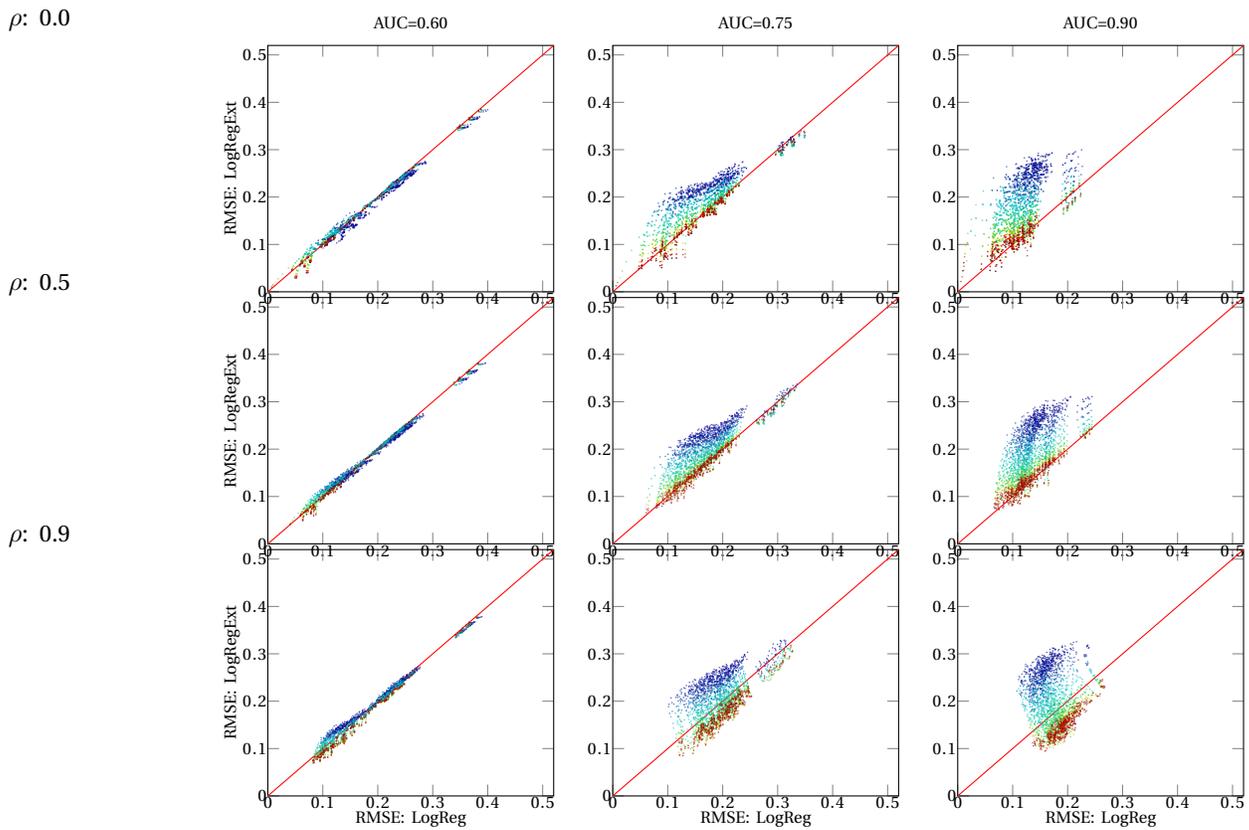

Figure 5b: multi-score calibration: RMSE$^{sub}$.

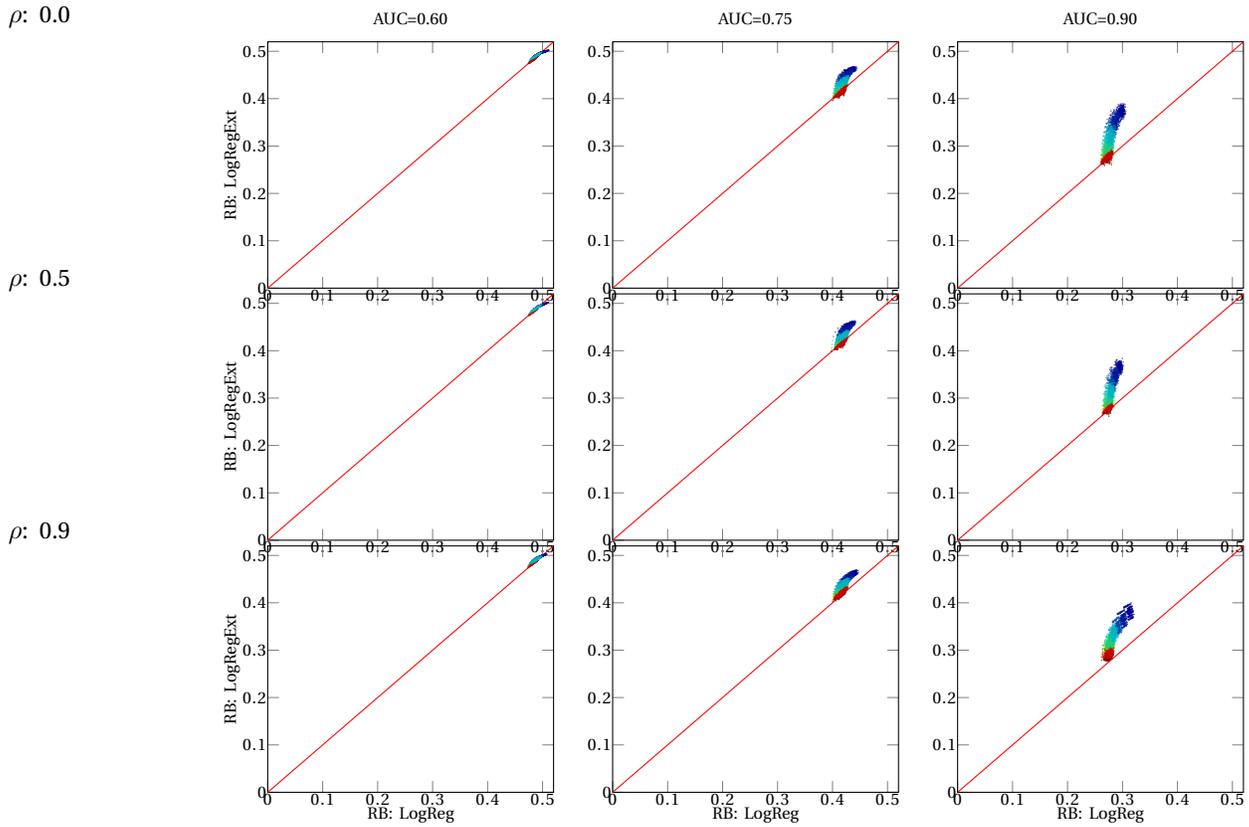

Figure 5c: multi-score calibration: RB$^{ind}$.

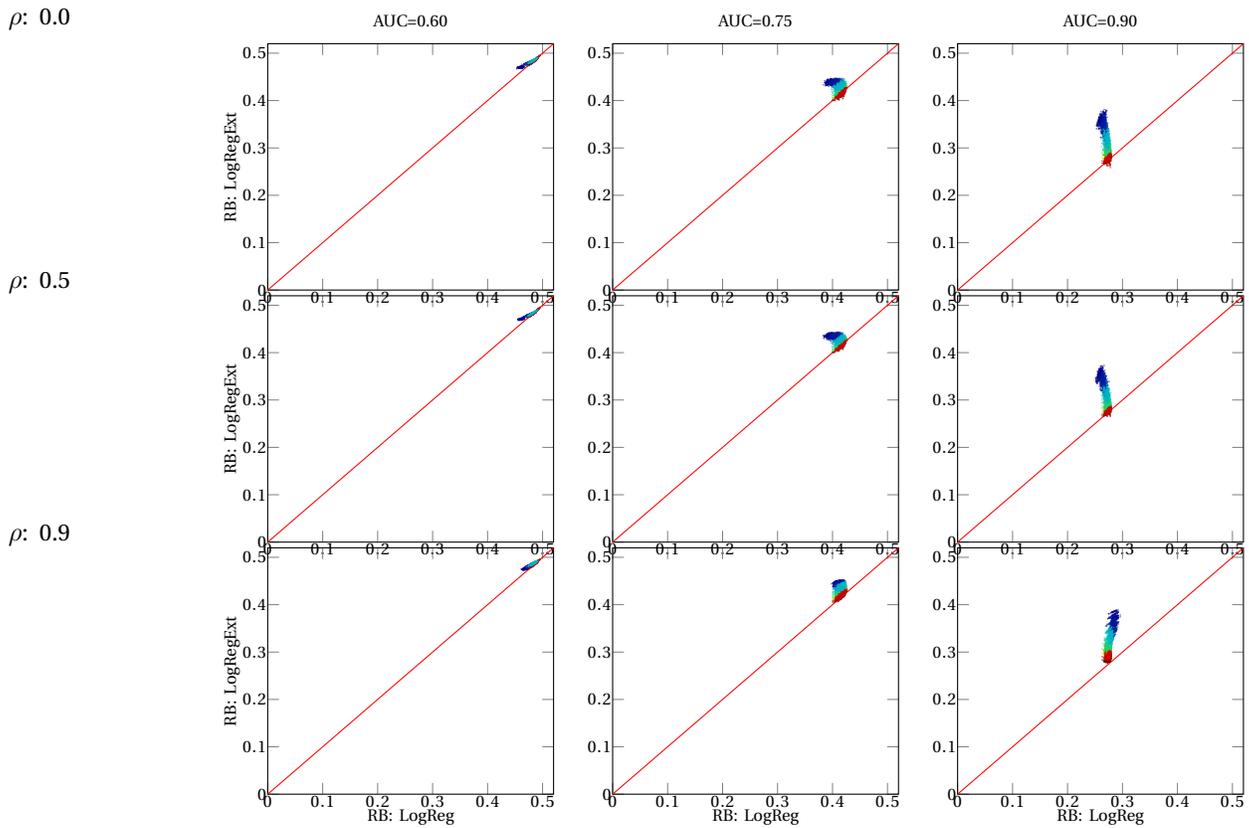

Figure 5d: multi-score calibration: RB$^{sub}$.

$\rho$: 0.0

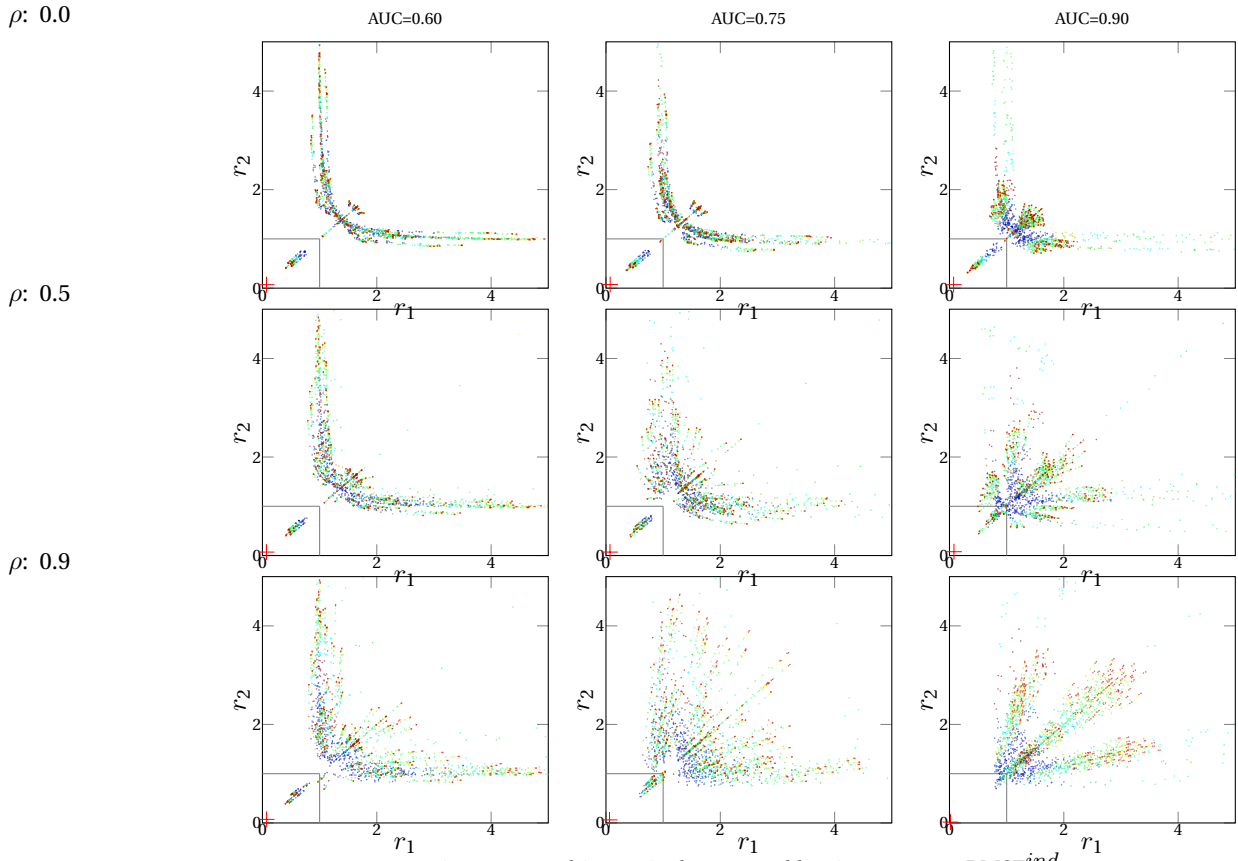

Figure 6a: multi- vs. single-score calibration, LogReg, RMSE$^{ind}$.

$\rho$: 0.0

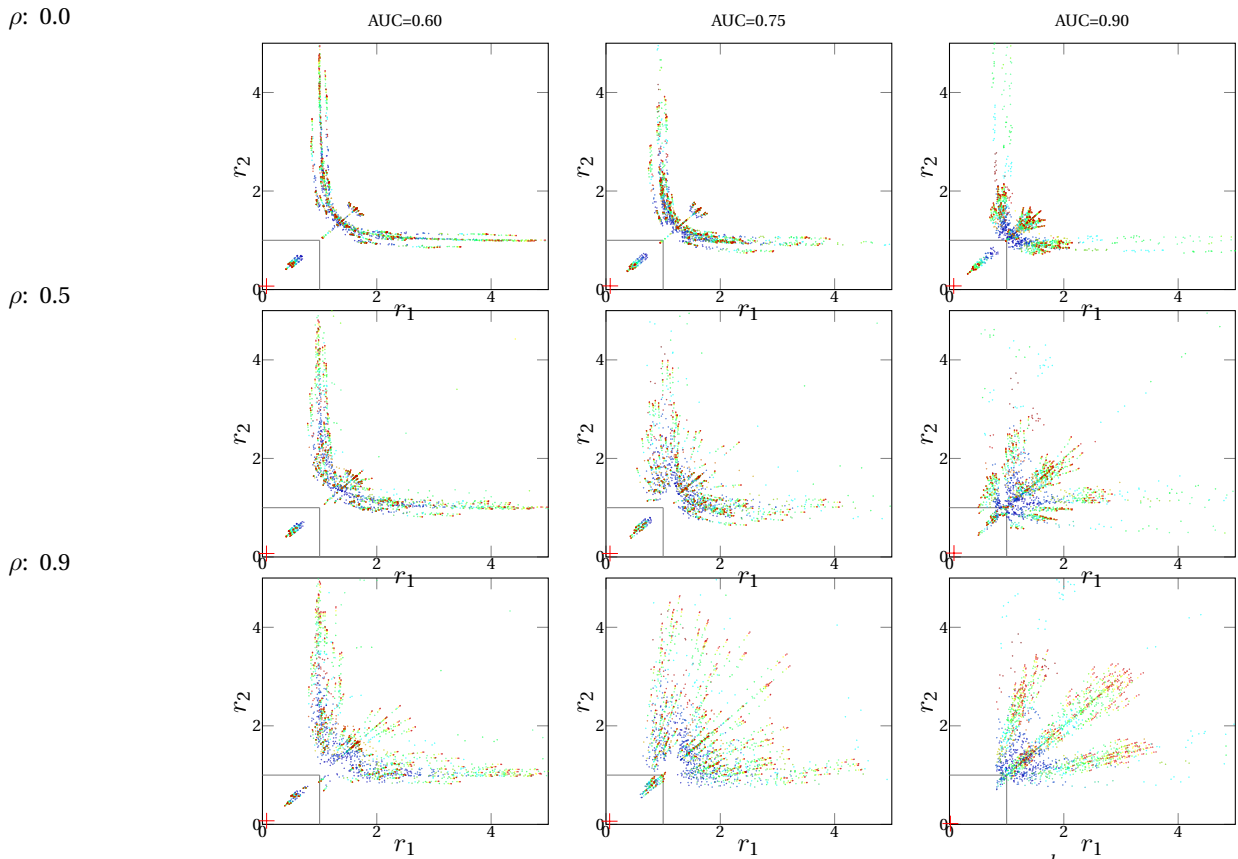

Figure 6b: multi- vs. single-score calibration, LogReg, RMSE$^{sub}$.

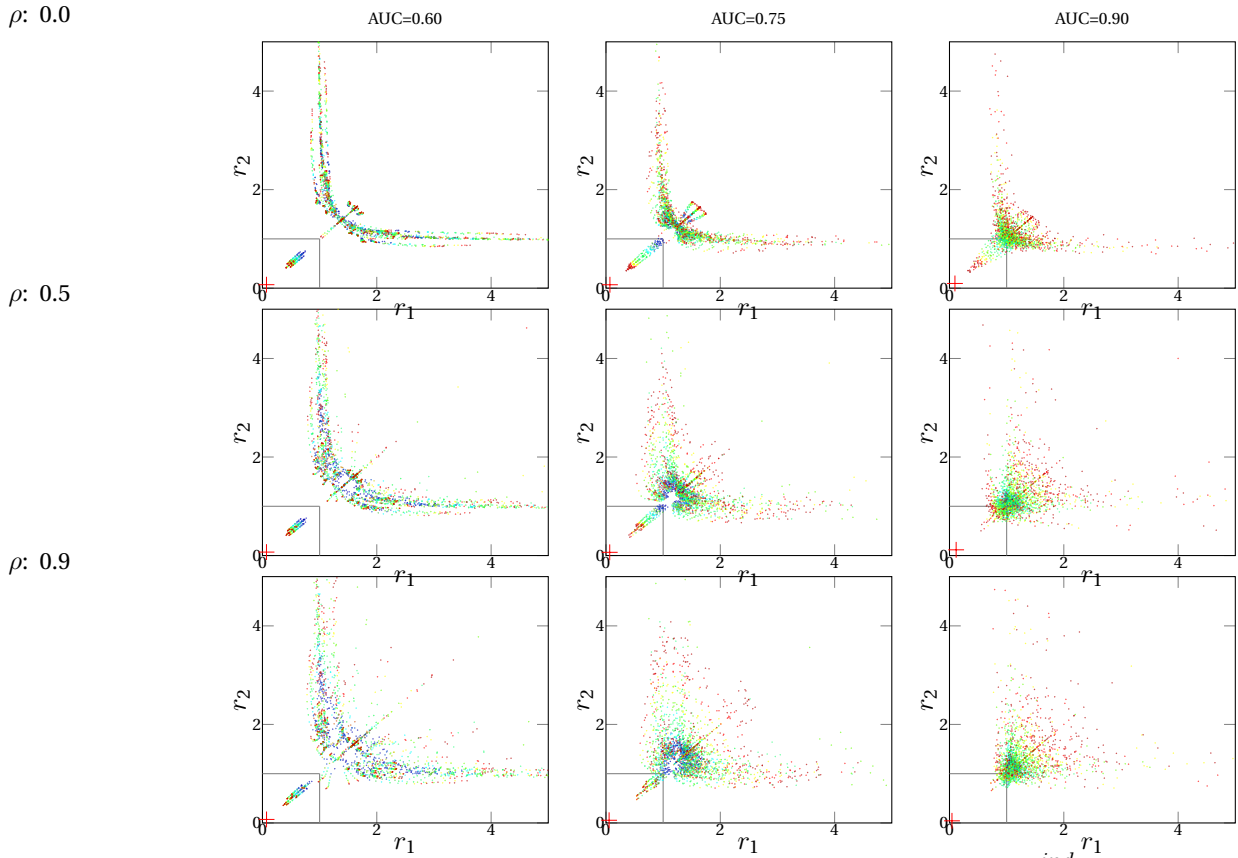

Figure 6e: multi- vs. single-score calibration, LogRegExt, RMSE$^{ind}$.

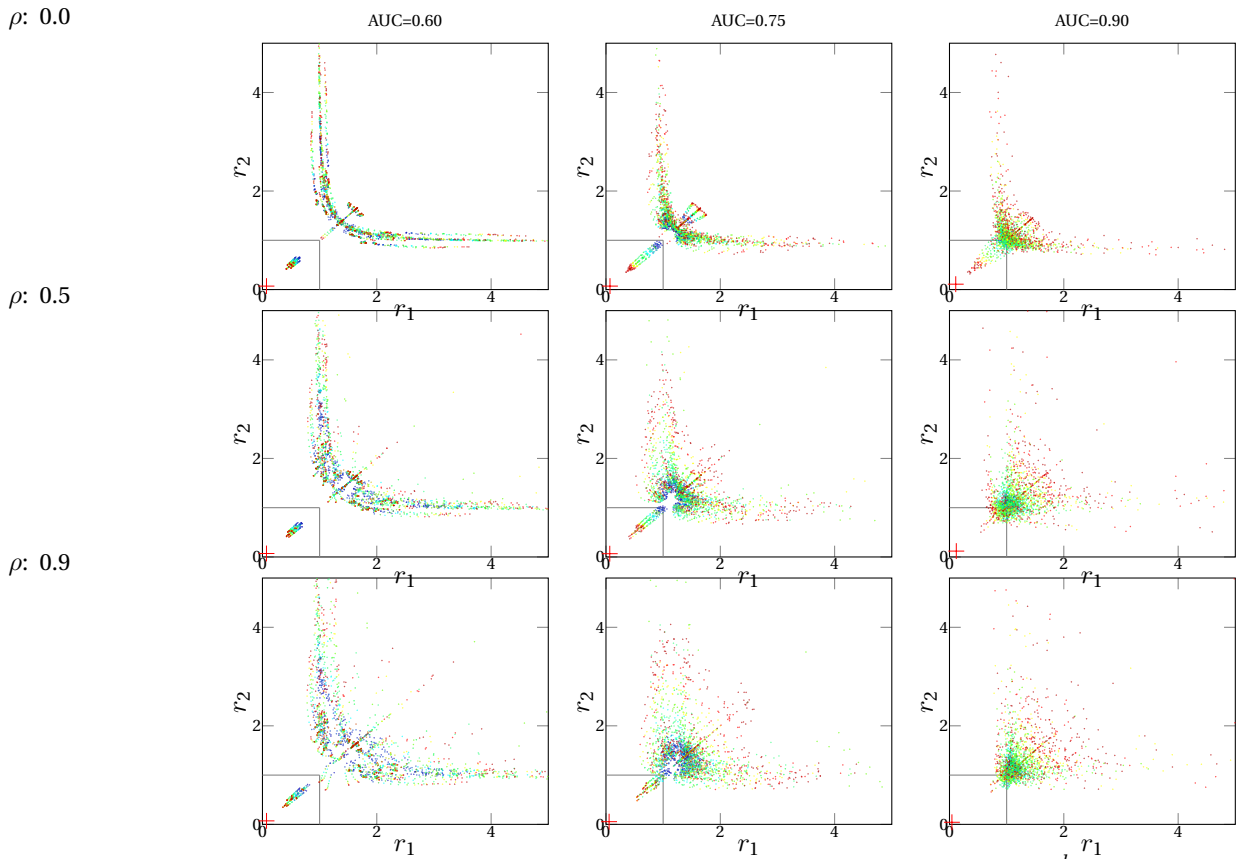

Figure 6f: multi- vs. single-score calibration, LogRegExt, RMSE$^{sub}$.

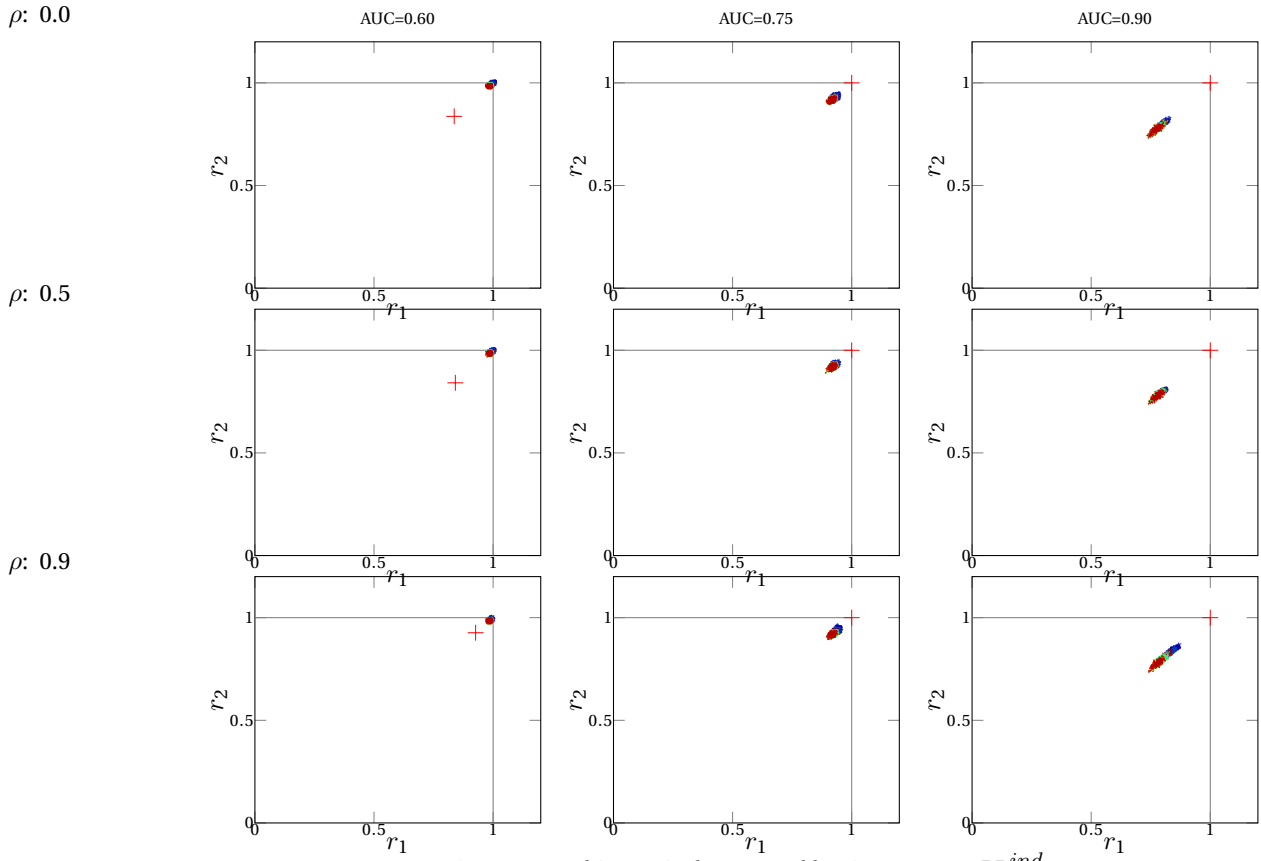

Figure 6c: multi- vs. single-score calibration, LogReg, RB$^{ind}$.

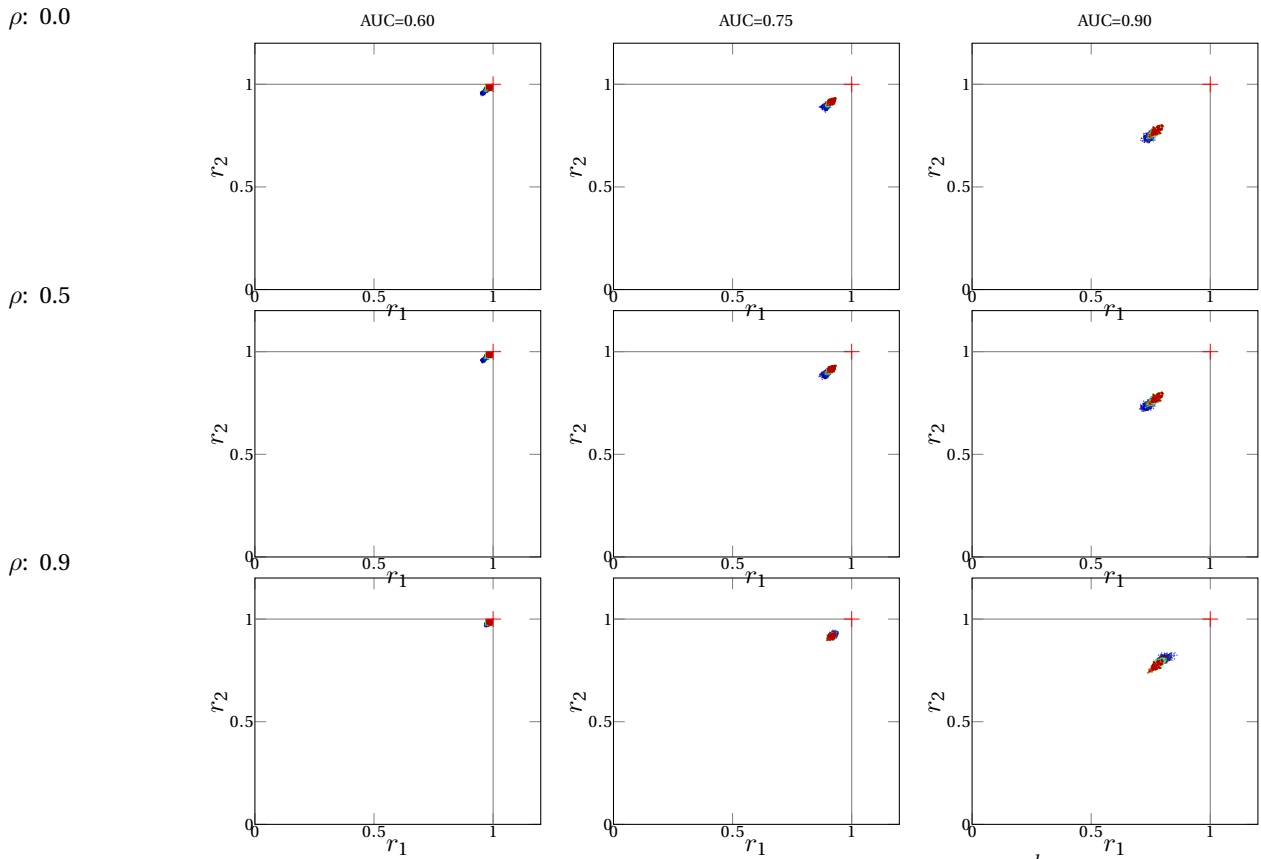

Figure 6d: multi- vs. single-score calibration, LogReg, RB$^{sub}$.

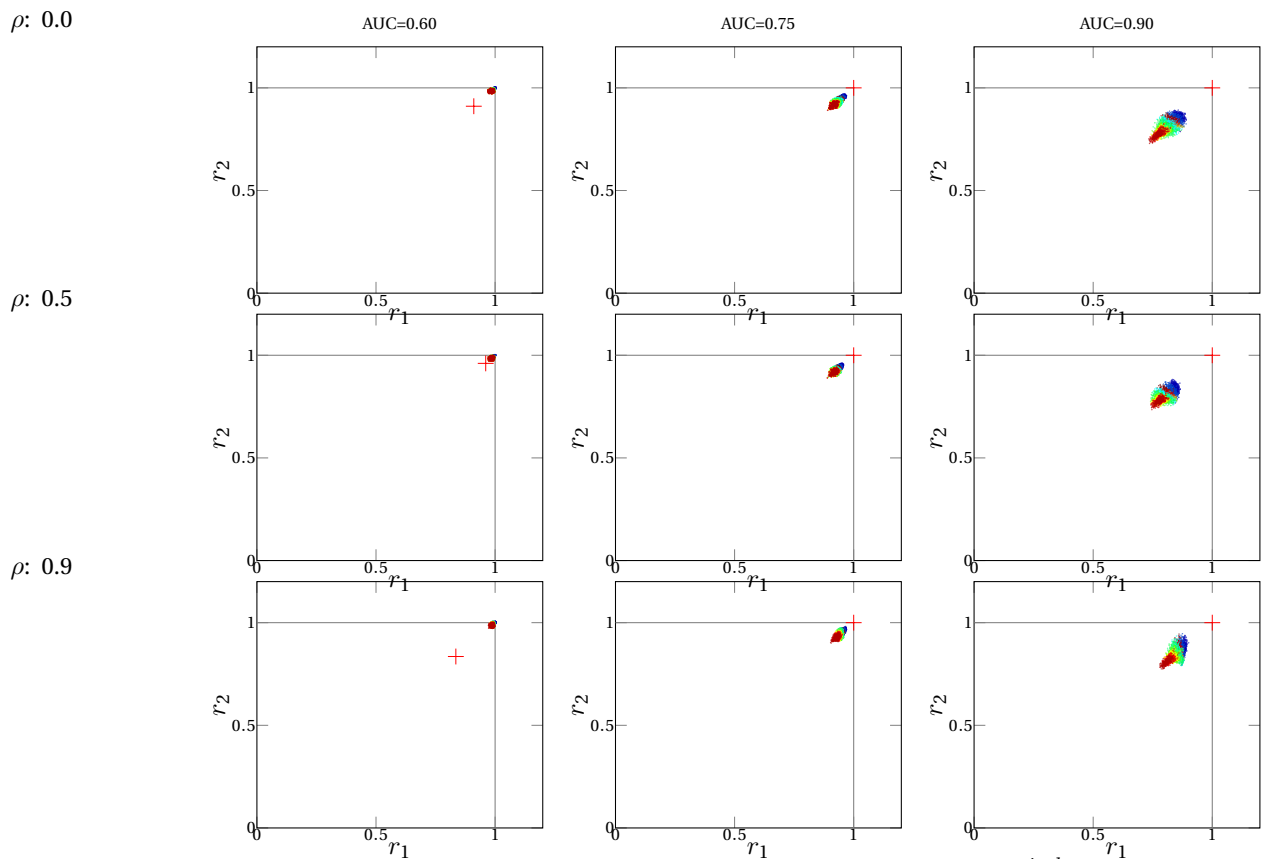

Figure 6g: multi- vs. single-score calibration, LogRegExt, RB$^{ind}$.

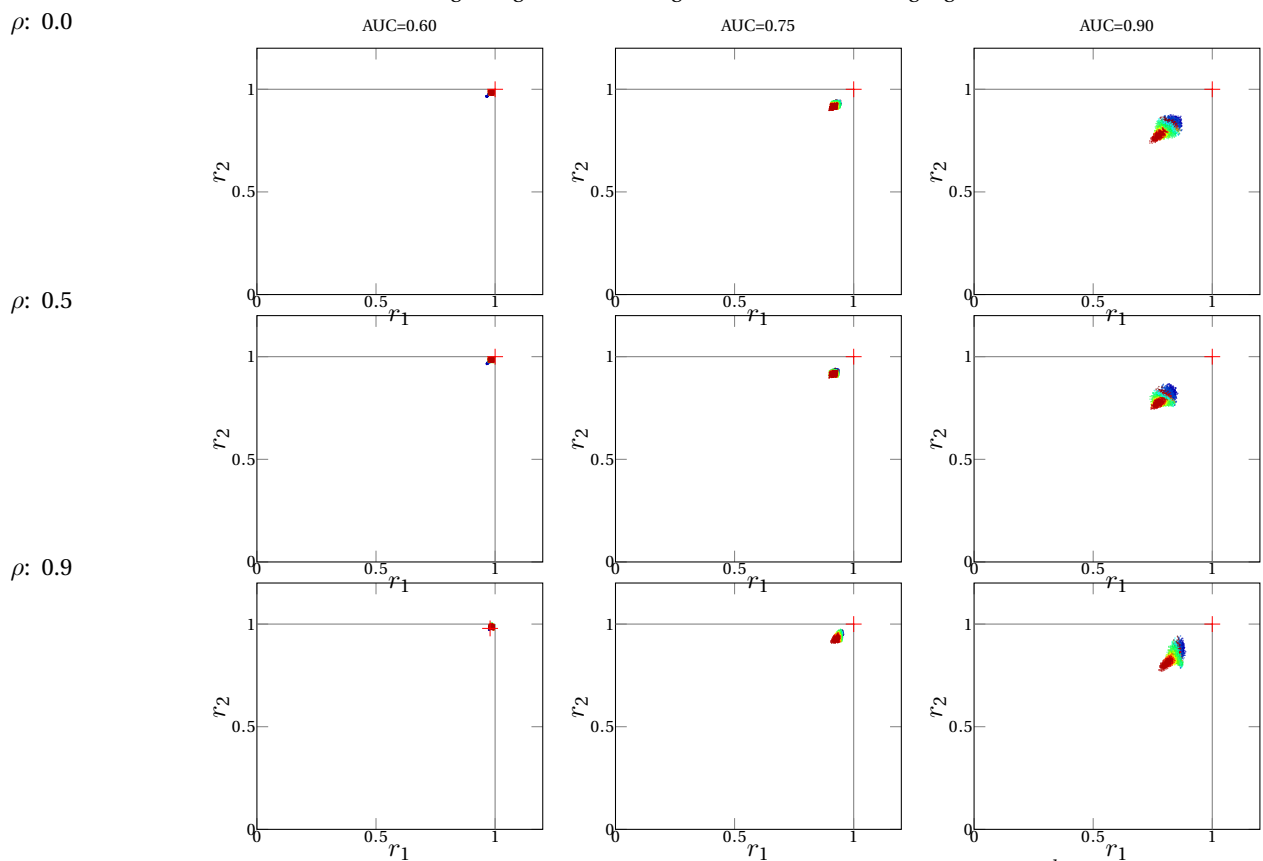

Figure 6h: multi- vs. single-score calibration, LogRegExt, RB$^{sub}$.